%% file: neurips_2024.tex
\documentclass{article}

    \PassOptionsToPackage{numbers, compress}{natbib}

    \usepackage[preprint]{neurips_2024}

\usepackage[utf8]{inputenc} 
\usepackage[T1]{fontenc}    
\usepackage{url}            
\usepackage{booktabs}       
\usepackage{amsfonts}       
\usepackage{nicefrac}       
\usepackage{microtype}      

\usepackage{amssymb}
\usepackage{bbm}

\usepackage{wrapfig}
\usepackage{caption}

\usepackage{tikz}
\usepackage{multicol}
\usepackage{multirow}
\usepackage{boxhandler}
\usepackage{tabularx}
\usepackage{adjustbox}
\usepackage{epstopdf}
\definecolor{mygray}{gray}{.9}
\definecolor{mygreen}{RGB}{46, 204, 113}
\definecolor{mydarkred}{RGB}{139,0,0} 
\definecolor{mydarkgreen}{RGB}{46,204,113} 

\newcommand{\red}[1]{\textcolor{red}{#1}}

\newcommand{\mygreen}[1]{\textcolor{blue}{#1}}

\usepackage{booktabs}
\input{math_commands.tex}

\usepackage{graphicx}
\graphicspath{{assets/}} 

\usepackage{lipsum}
\usepackage{makecell} 
\usepackage{subfigure}
\usepackage[pagebackref,breaklinks,colorlinks]{hyperref}
\usepackage{multirow}
\usepackage{enumitem}

\usepackage{colortbl}
\usepackage{pifont} 

\usepackage{xspace}
\makeatletter
\DeclareRobustCommand\onedot{\futurelet\@let@token\@onedot}
\def\@onedot{\ifx\@let@token.\else.\null\fi\xspace}

\def\eg{\emph{e.g}\onedot} 
\def\ie{\emph{i.e}\onedot} 
 
\def\etc{\emph{etc}\onedot}

\makeatother

\usepackage[pagebackref,breaklinks,colorlinks]{hyperref}
\hypersetup{
	colorlinks=true,
	linkcolor=red,
	filecolor=blue,      
	urlcolor=cyan,
	citecolor=green,
}

\title{\centering Assessing Model Generalization in Vicinity}

\author{%
  Yuchi Liu \\ 
  Australian National University\\
  \texttt{yuchi.liu@anu.edu.au} \\
  \And
   Yifan Sun \\
  Baidu \\
  \texttt{sunyf15@tsinghua.org.cn} \\
  \AND
  Jingdong Wang \\
  Baidu \\
  \texttt{wangjingdong@baidu.com} \\
  \And
  Liang Zheng \\
  Australian National University\\
  \texttt{liang.zheng@anu.edu.au} \\
}

\begin{document}

\maketitle


\begin{abstract}
This paper evaluates the generalization ability of classification models on out-of-distribution test sets without depending on ground truth labels. Common approaches often calculate an unsupervised metric related to a specific model property, like confidence or invariance, which correlates with out-of-distribution accuracy. However, these metrics are typically computed for each test sample individually, leading to potential issues caused by spurious model responses, such as overly high or low confidence.
To tackle this challenge, we propose incorporating responses from neighboring test samples into the correctness assessment of each individual sample. In essence, if a model consistently demonstrates high correctness scores for nearby samples, it increases the likelihood of correctly predicting the target sample, and vice versa. The resulting scores are then averaged across all test samples to provide a holistic indication of model accuracy. Developed under the vicinal risk formulation, this approach, named vicinal risk proxy (VRP), computes accuracy without relying on labels.
We show that applying the VRP method to existing generalization indicators, such as average confidence and effective invariance, consistently improves over these baselines both methodologically and experimentally. This yields a stronger correlation with model accuracy, especially on challenging out-of-distribution test sets. The code for our work is available \href{https://github.com/liuyvchi/Vicinal-Risk-Rroxy}{here}\footnote{\url{https://github.com/liuyvchi/Vicinal-Risk-Rroxy}}. 
\end{abstract}

\section{Introduction}

Because of the ubiquitous existence of distributional shifts in real-world systems, it is important to evaluate the generalization capacity of trained models on out-of-distribution (OOD) test data. 
In practical OOD scenarios, because obtaining test ground truths is expensive, model evaluation techniques that do not rely on test labeled are attracting increasing attention.

%

For this problem, unsupervised risk measurements are introduced that capture useful model properties, such as confidence and invariance. These measurements serve as indicators of a model's generalization ability. Importantly, for a sample of interest without ground-truth, these methods compute a risk proxy \textit{merely using this sample alone}. For example, \cite{hendrycks17baseline} use the maximum Softmax value of a test sample itself as its confidence score, and show that once averaged over the OOD test set, it serves as a reliable indicator. The Effective Invariance score \citep{deng2022strong} is computed as the prediction consistency between this sample and its transformed version (\eg, rotation and grey-scale). Because these indicators measure model properties that underpin its generalization ability, they generally exhibit fair correlation with the model's out-of-distribution accuracy.




\begin{figure*}
    \centering
    \captionsetup{type=figure}
    \includegraphics[width=1\textwidth]{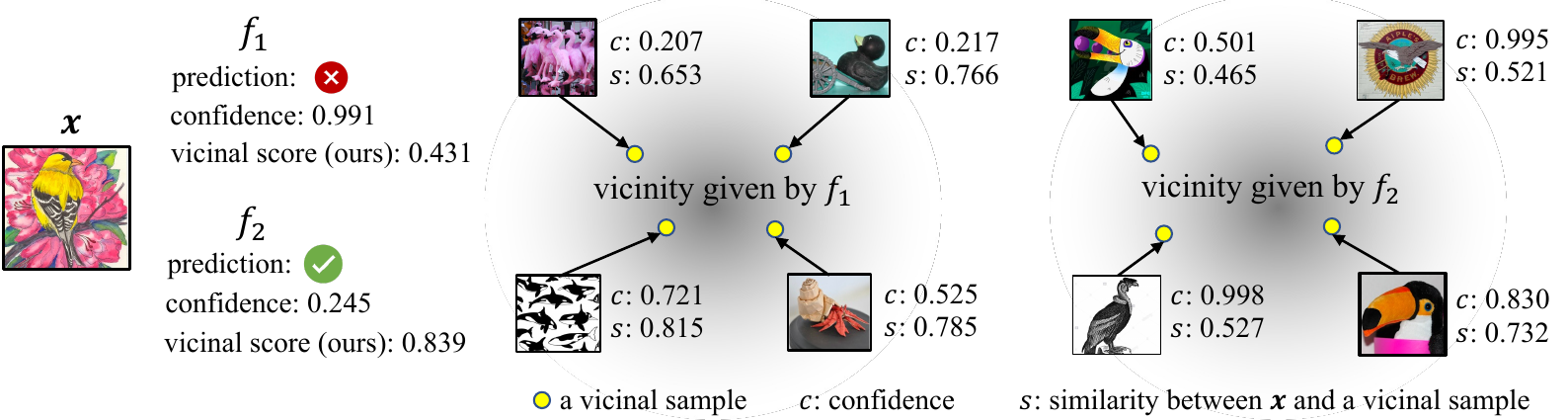}
    \vspace{-5mm}
    \captionof{figure}{\textbf{Illustrating spurious model responses and how our method corrects them, we use confidence as a model generalization indicator.} 
    For a given input $\bm{x}$ from ImageNet-R \citep{hendrycks2021many}, models $f_1$ and $f_2$ incorrectly and correctly classify it, respectively. However, the confidence score (0.991) for the incorrect prediction by classifier $f_1$ is excessively higher than the score (0.245) for the correct prediction by classifier $f_2$, indicating spuriousness. The model generalization ability ranking based on this test sample fails. Our proposed vicinal method, a similarity-weighted sum of confidence, provides more reasonable scores (0.431\protect\footnotemark and 0.839).
    %
    }
    \label{fig:first_figure_example}
    \vspace{-5mm}
\end{figure*}

However, we find this isolated way of measuring model effectiveness for a sample of interest suffers from spurious model responses. 
In Fig.~\ref{fig:first_figure_example}, we show two models where the one make incorrect prediction on a test sample may have a much higher confidence score than a model provide correct prediction. 
In other words, a high (\textit{resp.} low) confidence or invariance score sometimes does not mean a correct (\textit{resp.} incorrect) model prediction. 
These erroneous scores, once accumulated in the test set, would compromise the effectiveness of measuring (comparing) model risks.

To address this problem, when computing the risk proxy for a model on a test sample, we propose to integrate the risk proxy into the model behaviour on the adjacent samples, where such integration is performed using the vicinal distribution for the test sample (Fig. ~\ref{fig:first_figure_example}). Intuitively, if neighboring samples generally exhibit high risks (\emph{e.g.,} low confidence), the center sample with excessively low risk will be assigned an increased risk score, and vice versa. Here, the contribution of each neighboring sample to the center sample is proportional to their similarity. This strategy allows model responses (risk proxy score) to better indicate model prediction correctness for the test sample, as shown at the bottom of Fig. ~\ref{fig:model_histogram}. To indicate the overall generalization ability of models, we further compute the vicinal risk proxy (VRP) as the average individual vicinal score over the entire test set.

Another advantage of this vicinal assessment scheme is that it can be applied on top of various risk proxies based on individual test samples, including confidence, invariance and their variants. Our experiments show that VRP brings consistent improvements to them: stronger correlation between risk proxies rectified by VRP and model OOD accuracy over 200 different classifiers is generally observed on 9 benchmarks. 
In summary, this paper has the following main points.

\begin{itemize}[leftmargin=*]
    \item We examine existing methods in OOD generalization assessment in the lens of risk estimation. 
    \item We propose to integrate vicinal distribution of a sample into its risk estimate, to inhibit spurious model responses. 
    \item The proposed vicinal risk proxy (VPR), when applied to existing risk proxies, brings consistent improvement: stronger correlation is observed between vicinal proxies and model OOD accuracy. 
\end{itemize} 

\section{Related Work}

\footnotetext{We employ Eq.~\eqref{eq:vrp_approximation}. For instance, the vicinal score of $\bm{x}$ in Fig.~\ref{fig:first_figure_example} is computed as: {\scriptsize $0.431 \!=\! (0.653\!\times \!0.207\!+\!0.766\!\times\! 0.217\!+\!0.815\!\times\! 0.721\! +\! 0.785\! \times\! 0.525) / (0.653\!+\!0.766\!+\!0.815\!+\!0.785)$}.}

\textbf{Data-centric model generalization assessment}  aims to predict the accuracy of a \textit{given} model on \textit{various} unlabeled test sets. Average model confidence \citep{hendrycks17baseline, tu2023assessing} on the testing samples is a simple and useful indicator of model accuracy. 
\citet{guillory2021predicting} propose using the confidence discrepancy between the validation and test sets to correct the confidence-based accuracy. \citet{deng2021does} tackle this challenge by comparing models based on their accuracy in self-supervised tasks.
\citet{Garg2022Leveraging} predict accuracy by using the percentage of testing samples exceeding a threshold learned from a validation set in the source domain. In addition to confidence, domain shift can also be used as a cue to predict model accuracy on the target set \citep{guillory2021predicting,deng2021labels}. This paper does not focus on this setup and only provides some results in the supplementary materials. 

\textbf{Model-centric generalization assessment.} Some works focus on \textit{in-distribution} generalization \citep{garg2021ratt, jiang2021assessing, negrea2020defense, zhou2020uniform}. This paper instead studies \textit{OOD} generalization. In this problem, we train a variety of models on a training set and predict and compare their performance on an unlabeled \textit{OOD} test set. 
\citet{deng2022strong} propose effective invariance (EI) to measure the consistency between model predictions on the original image and its transformed versions. It is also feasible to use data-centric indicators such as average confidence \citep{hendrycks17baseline, tu2023assessing}. 
However, we find these methods sometimes give excessively high (low) scores to incorrect (correct) samples, which compromise performance assessment; we show this problem can be alleviated by the proposed method. 

\textbf{Vicinal risk}
was originally introduced in the vicinal risk minimization (VRM) principle \citep{chapelle2000vicinal}. In VRM, each training sample defines a vicinal distribution, and accordingly, model risk is evaluated based on these distributions instead of individual training samples. VRM is widely reflected in data augmentation methods 
\citep{chapelle2000vicinal, cao2015use, ni2015training, zhang2017mixup, yun2019cutmix, qin2020resizemix, kim2020puzzle}. For example, MixUp \citep{zhang2017mixup} generates samples from vicinal distributions by 
mixing two images and their corresponding labels with random mixing weights. Other examples are 
CutMix \citep{yun2019cutmix}, ResizeMix \citep{qin2020resizemix}, and PuzzleMix \citep{kim2020puzzle}.
While these methods reflect vicinal risks on labeled training data, we apply this idea to unlabeled test data. Our strategy smooths out spurious proxy scores and allows for better approximation to model out-of-distribution accuracy.

\section{Preliminaries}
\subsection{Risk and Accuracy in Supervised Evaluation} 

We consider a model $f$ belonging to a class of models $\mathcal{F}$ and a target distribution $P(\bm{x},y)$. Model risk can be formulated as the expectation of a given loss function $\ell(f(\bm{x}), y)$ on the distribution $P$ during the test stage:
\begin{equation}
R(f)=\int\ell(f(\bm{x}), y)dP(\bm{x},y).
\label{eq:risk_exp}
\end{equation}
In practice, since distribution $P(\bm{x}, y)$ is unknown,  Eq. \eqref{eq:risk_exp} cannot be directly computed. Standard practice thus  
approximates the test risk by replacing $P(\bm{x}, y)$ with an empirical distribution $P_{emp}(\bm{x}, y)$, formed by assembling Dirac delta functions~\citep{dirac1981principles} centered at each sample in a given test set $\mathcal{D}:=\{(x_i, y_i)\}_{i=1}^{n}$:
\begin{equation}
dP_{emp}(\bm{x}, y)=\frac{1}{n}\sum_{i=1}^{n}\delta_{\bm{x}_i}(\bm{x})\cdot\delta_{y_i}(y).
\label{eq:emoirical_distribnution}
\end{equation}
Substituting Eq. \eqref{eq:emoirical_distribnution} into Eq. \eqref{eq:risk_exp}, the empirical risk under empirical target distribution $P_{emp}$ becomes:  
\begin{equation}
    R_{emp}(f)=\frac{1}{n}\sum_{i=1}^{n}\ell(f(\bm{x}_i), y_i).
    \label{eq:emp_risk}
\end{equation}
By convention, accuracy can be viewed as a type of empirical risk with the accuracy loss: $\ell_{acc}= 0, \mbox{if }\hat{y}=y,  \mbox{otherwise} \ell_{acc}=1$ 
where $\hat{y}$ is the predicted class with maximum (Softmax) confidence in $f(\bm{x})$.

\subsection{Vicinal Risk Minimization}
\label{Sec:VRM}
In \textit{training}, the risk (loss values) of individual \textit{training} samples may not fairly reflect the true generalization ability of a model on this sample. That is, the model may trivially minimize $R_{emp}(f)$ in Eq. \eqref{eq:emp_risk} by mere memorization of the samples \citep{zhang2017mixup, zhang2021understanding} rather than learn effective patterns.
To address this problem, 
\textit{vicinal risk minimization} ~\citep{chapelle2000vicinal} suggests to replace the Dirac delta function $\delta_{\bm{x}_i}(\bm{x})$ and $\delta_{y_i}(y)$ in Eq. \eqref{eq:emoirical_distribnution} by some density estimates of the vicinity of point $(\bm{x}_i, y_i)$:
\begin{equation}
dP_v(\bm{x},y)=\frac{1}{n}\sum_{i=1}^{n}dv(\bm{x},y | \bm{x}_i, y_i),
\label{eq:vicinal_distrinution_mixture}
\end{equation}
where $dv(\bm{x},y| \bm{x_i}, y_i)$ is the vicinal density function describing the probability of finding point $(\bm{x},y)$ in the vicinity of $(\bm{x}_i, y_i)$, and $P_v$ is a mixture distribution of $n$ vicinal distributions $v$. The expectation of the vicinal risk of model $f$ is now the mean of risk expectation in each vicinal distribution $v$:
\begin{align}
\begin{split}
    R_{v}(f) &= \int \ell(f(\bm{x}), y) dP_v(\bm{x},y) = \frac{1}{n}\sum_{i=1}^{n}\int\ell(f(\bm{x}), y_i)dv(\bm{x},y | \bm{x}_i,y_i).
\end{split}
\label{eq:vicinal_risk}
\end{align}
The uniform vicinal distribution~\citep{chapelle2000vicinal}, Gaussian vicinal distribution~\citep{chapelle2000vicinal} and mixup vicinal distribution~\citep{zhang2017mixup} are well-known vicinity in the risk minimization task~\citep{zhang2018generalization}. The success of vinical risk minimization in \textit{model training} \citep{chapelle2000vicinal, cao2015use, ni2015training, hai2010vicinal, dong2022revisiting, zhang2017mixup} demonstrates the effectiveness of this idea, which 
inspires us to enhance the risk estimation in the unsupervised \emph{evaluation} problem.

\section{Methodology}

\begin{figure*}[t]%
\centering
\includegraphics[width=\textwidth]{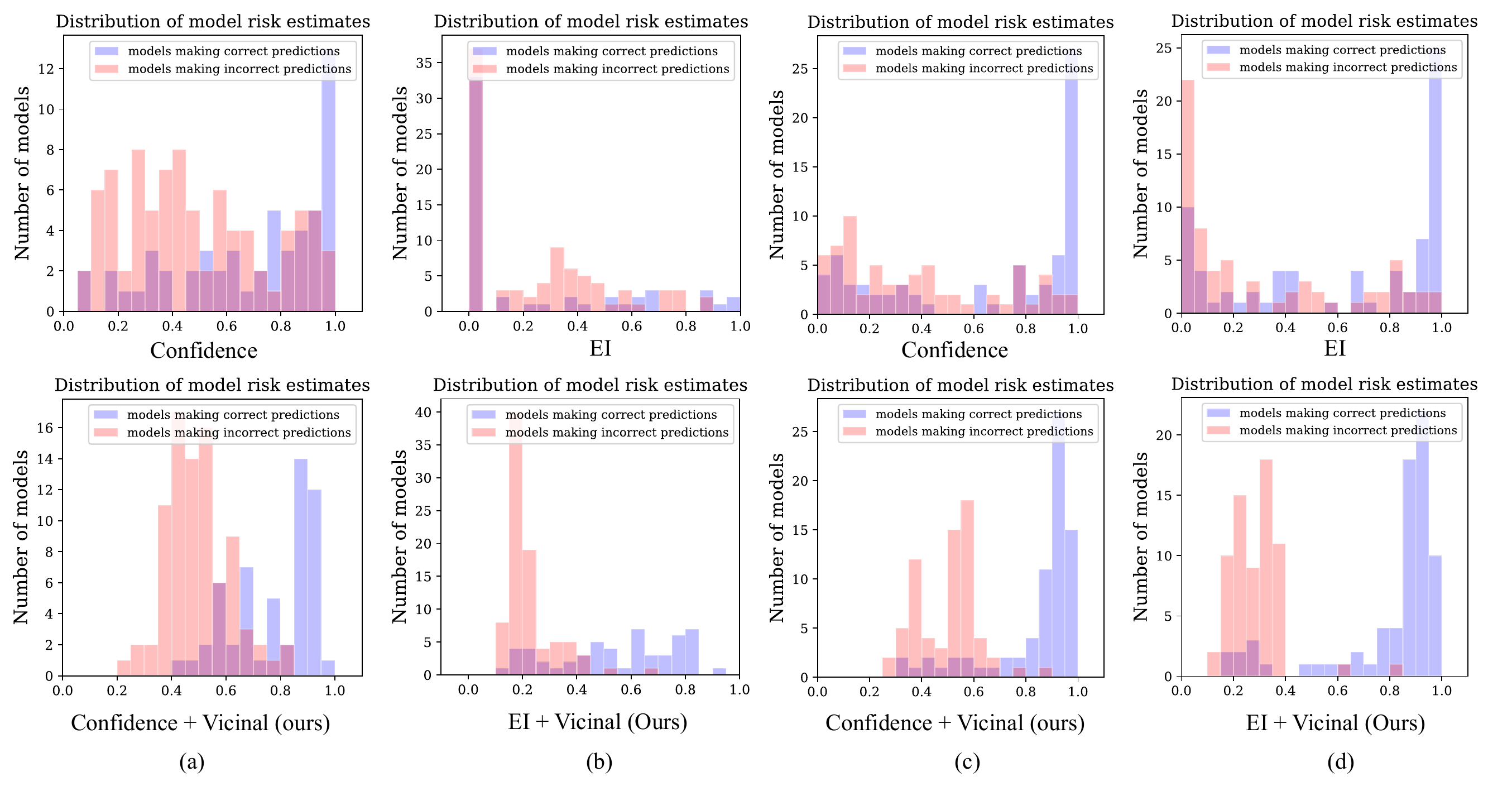}
\vspace{-5mm}
\caption{\textbf{Examples demonstrating how vicinal risks of individual samples can more efficiently differentiate between models making correct and incorrect predictions are provided.}
In (a) - (b), a single test sample from ImageNet-R is used along with 140 models trained on the ImageNet training set. We employ risk estimates based on confidence\citep{tu2023assessing, hendrycks17baseline}, confidence combined with our method, EI \citep{deng2022strong}, and EI combined with our method. The distributions of these risk estimates across the 140 models for a given test sample are illustrated.
In (c) - (d),   results are reported for another test sample. Notably, confidence and EI, which rely on the sample in isolation, lead to spurious model responses. In the top row, many models making incorrect predictions may exhibit excessively high confidence/EI, while those making correct predictions may show unexpectedly low confidence/EI. In contrast, our method (bottom row)) effectively corrects these erroneous risk estimates, enhancing the separation of risk estimates on individual samples between good and poor models. Consequently, the vicinal risk proxy averaged over the entire out-of-distribution (OOD) test set becomes a more reliable indicator of model accuracy. \textbf{Additional examples are provided in the supplementary material (Fig.~\ref{fig:model_histogram_AC},~\ref{fig:model_histogram_DoC} and~\ref{fig:model_histogram_ATC}), and further statistical results are presented in Fig \ref{fig:overlap_bar}.}}
\label{fig:model_histogram}
\vspace{-0.3cm}
\end{figure*}

\subsection{Risk Proxy in Unsupervised Evaluation}
\label{Sec:risk_proxy}

In unsupervised evaluation, it is infeasible to use $\ell_{acc}$ defined in Eq. \eqref{eq:emp_risk} 
because of the absence of ground-truths. To still be able to indicate model risk, existing methods typically 
design a \emph{proxy} loss $\hat{\ell}$ and compute its expectation on the unlabeled distribution $P(\bm{x})$: 
\begin{equation}
    \widehat{R(f)}=\int\hat{\ell}(f, \bm{x}, \varphi)dP(\bm{\bm{x}}),
    \label{eq:risk_proxy_exp}
\end{equation}
where $\widehat{R(f)}$ is defined as \textit{risk proxy}. $\hat{\ell}$, usually reflecting crucial model properties (\eg, confidence and invariance), is computed based on the response of model $f$ to input $\bm{x}$, and additional  knowledge $\varphi$ of the model. 
In average confidence \citep{tu2023assessing} and EI \citep{deng2022strong}, $\hat{\ell}$ takes the confidence of $\bm{x}$ or its transformation, so $\varphi=\emptyset$.  In DoC \citep{guillory2021predicting},  $\varphi$ means model accuracy and average confidence evaluated on a validation set. In ATC \citep{Garg2022Leveraging}, $\varphi$ is a model-specific confidence threshold learned from a validation set.

In practice, on a test set with $n$ test samples, $\widehat{R(f)}$ is approximated by the \textit{empirical risk proxy}:
\begin{equation}
\widehat{R_{emp}(f)}=\frac{1}{n}\sum_{i=1}^{n}\hat{\ell}(f, \bm{x}_i, \varphi).
\label{eq:emp_risk_proxy}
\end{equation}

\subsection{Proposed Vicinal Risk Proxy}\label{sec:vicinal_risk}

\textbf{Issues of empirical risk proxies.} 
Existing methods in unsupervised evaluation assume that the designed risk proxy $\widehat{R(f)}$ 
well correlates with model generalization on the target OOD distribution. However, this assumption can be tenuous when we zoom in individual samples. As depicted at the top of Fig. \ref{fig:model_histogram}, many models correctly classifying a sample have unexpected low confidence/invariance scores, and many incorrectly model predictions have excessively high confidence/invariance scores. The presence of such spurious model responses in individual samples introduces noise to the empirical risk proxy defined in Eq. \eqref{eq:emp_risk_proxy}, rendering it less effective in assessing model generalization capabilities. The dual problem in \textit{training} is described in Section \ref{Sec:VRM}.



\textbf{Solution.} 
Given an unlabeled test set $\mathcal{D}:=\{(\bm{x}_i)\}_{i=1}^{n}$, we propose to compute the \textit{vicinal risk proxy} on $\mathcal{D}$ as an unsupervised indicator of the accuracy of model $f$ on this test set. 
We first define vicinal distribution $\mu$ for each test sample $\bm{x}_i$ below:
\begin{equation}
\begin{split}
\mu(\bm{x},y|f, \bm{x}_i).
\end{split}
\label{eq:vicinal_distribution}
\end{equation}
The probability density function for $\mu$ is defined as:
\begin{equation}
\begin{split}
d\mu(\bm{x},y|f, \bm{x}_i)= 
\begin{cases}
s(f(\bm{x}^t), f(\bm{x}^t_i)), & \mbox{if } \widehat{y^t}=\widehat{y^t_i} \\
0, & \mbox{if } \widehat{y^t}\neq\widehat{y^t_i}
\end{cases},
\end{split}
\label{eq:vicinal_density_f}
\end{equation}
where $\bm{x}^t$ is a transformed view of $\bm{x}$, and $\widehat{y^t_i}$ is the predicted class of $\bm{x}^t_i$. 
There are various options for image transformations in practice (\eg, grayscale, color jitter, rotation, \etc), and we have empirically selected rotation for simplicity.
The function $s(\cdot, \cdot)$ calculates the similarity between outputs generated by the model $f$, with the similarity computed empirically through the dot product\footnote{We do not constrain the choice of the similarity metric in our method; however, for simplicity in our experiments, we opt for the dot product.}. of the Softmax vectors.
Intuitively, $\mu$ is the probability distribution of finding pair $(\bm{x}, y)$ in the vicinity of $\bm{x}_i$, and $d\mu$ is its probability density function. Integrating such vicinal assessment into the point-wise empirical risk proxy, Eq. \eqref{eq:emp_risk_proxy} 
can be updated as the vicinal risk proxy:
\begin{equation}
\widehat{R_{v}(f)}=\frac{1}{n}\sum_{i=1}^{n} \int \hat{\ell}(f, \bm{x}, \varphi)d\mu(\bm{x},y|f, \bm{x}_i).
\label{eq:vicinal_risk_proxy}
\end{equation}
Essentially, instead of merely using $\bm{x}_i$ itself for risk estimation, we also use its neighboring samples. A sample with higher similarity with $\bm{x}_i$ contributes more to the risk. We find that spurious model responses on $\bm{x}_i$ can be effectively inhibited by its vicinal risks. For example, at the bottom of Fig. \ref{fig:model_histogram}, for a test sample, models making correct and incorrect predictions are better separated. Quantitative analysis will be provided in Section \ref{sec:exp}.

In practice, we approximate the expectation of $\hat{\ell}$ within the $i$-th distribution $\mu$ as:
\begin{equation}
\mathbb{E}_i= \frac{ \sum_{j=1}^{m_i} \hat{\ell}(f, \bm{x}_j, \varphi)d\mu(\bm{x}_j, \widehat{y_j}|f, \bm{x}_i)}{\sum_{j=1}^{m_i}d\mu(\bm{x}_j, \widehat{y_j}|f, \bm{x}_i)},
\label{eq:vrp_approximation}
\end{equation}
where $\widehat{y_j}$ is the predicted class of $\bm{x}_j$ in $\mathcal{D}$, and $m_i$ is the number of vicinal samples in a vicinal distribution of $\bm{x}_i$\footnote{
By default, $m_i$ is the count of samples with a probability density value greater than 0 in Eq. \eqref{eq:vicinal_density_f}. 
Experiments illustrating the impact of $m_i$ are presented in Fig.~\ref{fig:neighbors_num}.
}.
Intuitively, Eq. \eqref{eq:vrp_approximation} gives an empirical estimation of the risk proxy considering the vicinal distribution of $\bm{x}_i$ and the probability density defined in Eq. \eqref{eq:vicinal_density_f} for each vicinal sample. 
We provide the computational cost analysis for our method in Appendix~\ref{sec:app-complex}.

For individual samples, the vicinal score allows correct and incorrect model predictions to be better separated (refer Fig. \ref{fig:model_histogram} for an example). Collectively on the test set, models making more correct predictions (higher accuracy) will receive higher vicinal scores than models make less correct prediction (lower accuracy).
An illustrative derivation is presented in Appendix for further clarity.



\subsection{Apply Vicinal Risk Proxy to Existing Proxies.}

In Eq. \eqref{eq:risk_proxy_exp} and Eq. \eqref{eq:emp_risk_proxy}, we show that some existing approaches in unsupervised evaluation can be seen as unsupervised proxies for the empirical risk on the test set. 
Moreover, Eq. \eqref{eq:vicinal_risk_proxy} means that the proposed vicinal risk proxy marries existing risk proxies with vicinal distribution. In other words, the idea of considering vicinal samples can be applied to various proxy loss functions $\hat{\ell}$. For example, when $\hat{\ell}$ is sample confidence, or equivalently, $\widehat{R_{emp}(f)}$ is the test average confidence \citep{tu2023assessing, hendrycks17baseline} computed empirically, also called empirical risk proxy (ERP). $\widehat{R_v(f)}$ is the the vicinal average confidence under our vicinal assessment, so is called vicinal risk proxy (VRP). By default, we search for neighboring samples for each vicinal distribution throughout the entire dataset. Samples with similarities greater than $0$ are used to approximate the VRP score in Eq. \eqref{eq:vrp_approximation}. for this vicinity. 

\subsection{Discussions}
\textbf{Why do spurious model responses occur? 
}
Spurious model responses can occur for various reasons. One prominent factor is the over-confidence/under-confidence problem, as highlighted by \citet{guo2017calibration} in their work on calibration. 
While model calibration can assist in alleviating this issue in in-distribution (IND) datasets, spurious model responses become inevitable in out-of-distribution (OOD) scenarios. Consequently, exploring effective solutions to assess model generalization while considering the negative impact of these spurious responses is an intriguing area.

\textbf{Effectiveness of vicinal assessment under in-distribution test sets.} For IND data, models generally have consistently good behaviours on test samples (including their neighbors, apparently). As such, considering the vicinity does not give additional knowledge. In fact, we empirically find that applying our method on IND data does not compromise the system (refer Table \ref{tab:benchmark_testing}). Therefore, the proposed vicinal assessment is safe to use for both IND and OOD environments. 

\textbf{How can we guarantee neighborhood samples give correct responses?} We do not assume this. Instead, we look at response consistency in the neighborhood. If the neighborhood responses are consistent with the center sample, the risk of the center sample is reliable; if responses are incorrect, the risk of the center sample would not be reliable and needs to be balanced by its vicinity responses. 


\textbf{Vicinal assessment for data-centric unsupervised evaluation.} When we assume fixed classifier and training data and vary the test data \citep{deng2021does, Garg2022Leveraging}, technically vicinal assessment can be applied. 
However, the vicinal score of a single sample is only demonstrated to separate different \textit{models} (see Fig. \ref{fig:model_histogram}). In the date-centric scenario, it is not even possible to let a single sample separate different datasets, simply because different datasets do not share image samples. As such, we report that vicinal risk does not have noticeable improvement under the data-centric setup in the supplementary material.


\textbf{Application to other risk proxies.} The risk proxies considered in this paper, \eg, EI \cite{deng2022strong} and AC \cite{tu2023assessing, hendrycks17baseline}, are computed for individual samples. For model generalization analysis, some existing works are computed from the entire test set, such as ProjNorm~\cite{pmlr-v162-yu22i}, Dispersion Score~\cite{xie2023importance} and COT~\cite{lu2023characterizing}. We speculate that our method cannot be applied to these global scoring methods. 


\section{Experiments}
\label{sec:exp}

\begin{figure*}[t]
\centering
\includegraphics[width=\textwidth]{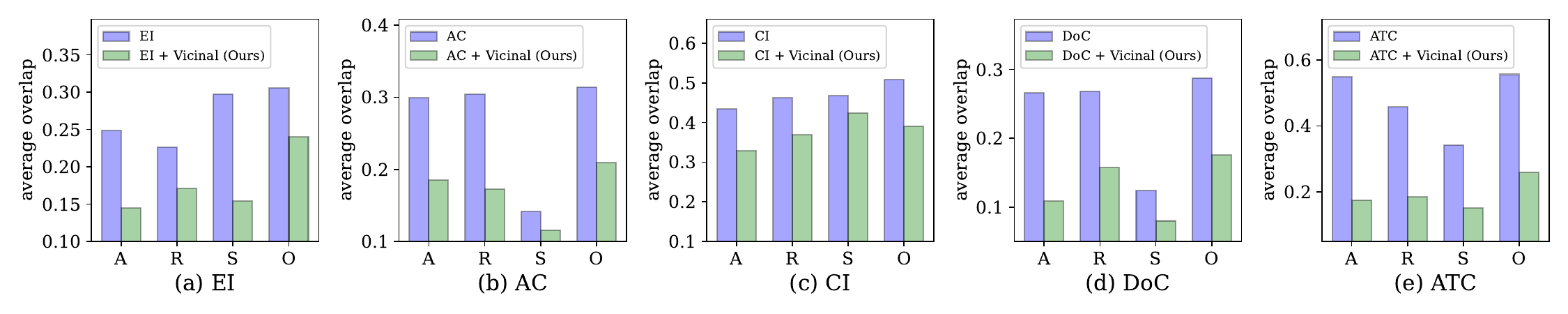}
\vspace{-4mm}
\caption{\textbf
{The average overlap of risk estimates for individual samples between correct and incorrect model predictions.
} We first estimate the distributions of risk estimate scores for correct and incorrect model predictions (140 in total) for each test sample. Then, the overlap of the two distributions for each sample is computed and finally averaged over the entire test set.   
All models are trained on ImageNet. In each figure, we use four test sets, ImageNet-A (A), ImageNet-R (R), ImageNet-S (S), and ObjectNet (O). From (a) to (e), EI, AC, CI, DoC, and ATC are used as baselines, respectively. 
\textit{a smaller value indicates lower overlap or higher separability}. 
We clearly observe that vicinal risk  scores (ours) statistically better differentiate models making correct and incorrect predictions by better separating their scores.} 
\label{fig:overlap_bar}
\vspace{-0.3cm}
\end{figure*}

\subsection{Datasets and Evaluation Metrics}
\label{Sec:Benchmark}

\textbf{ImageNet-1k setup.}
\underline{1. Model.} We use 140 models that have been trained or fine-tuned using the ImageNet-1k~\citep{deng2009imagenet} training set. We source these models from the models zoo Timm~\citep{rw2019timm}. As suggested by \citet{deng2022strong}, these models exhibit a diverse range of architectures, 
training strategies, and pre-training settings.
\underline{2. Data.} 
(1) \emph{ImageNet-A(dversarial)}~\citep{hendrycks2021nae} comprises natural adversarial examples that are unmodified and occur in the real-world. 
(2) \emph{ImageNet-S(ketch)}~\citep{wang2019learning} contains images with a sketch-like style.
(3) \emph{ImageNet-R(endition)}~\citep{hendrycks2021many} comprises of 30,000 images that exhibit diverse styles.
(4) \emph{ImageNet-Blur}~\citep{hendrycks2019benchmarking} was produced by applying a Gaussian function to blur the images from ImageNet-Val. We use blur with highest severity. 
(5) \emph{ObjectNet} \citep{barbu2019objectnet} is a real-world set for object recognition with control where object backgrounds, rotations, and imaging viewpoints are random.
(6) \emph{ImageNet-V2} \citep{recht2019imagenet} is a reproduced ImageNet dataset, whose distribution is similar to ImageNet.

\textbf{CIFAR10 setup.}
\underline{1. Model.} 
We use 101 models in this set sup. We follow \citet{deng2022strong} to access model weights. 
\underline{2. Data.} 
(1) \textit{CINIC-10} \citep{darlow2018cinic} is a fusion of CIFAR-10 and ImageNet-C \citep{hendrycks2019benchmarking} image classification datasets. It contains the same 10 classes as CIFAR-10. (2) \textit{CIFAR-10.1} \citep{recht2018cifar10.1} is produced with almost the same distribution as CIFAR-10. 

\textbf{iWildCam setup.} 
\underline{1. Model.} 
We use 35 models trained on the iWildCam\citep{beery2020iwildcam} training set.
\underline{2. Data.}  iWildCam-OOD test set contains animal pictures captured
in the wild.

\textbf{Evaluation metrics.}
We use the same evaluation metrics as \citep{deng2022strong}, \emph{i.e.}, Pearson's Correlation coefficient ($\gamma$)~\citep{cohen2009pearson} and Spearman’s Rank Correlation coefficient ($\rho$)~\citep{kendall1948rank}. They assess the degree of linearity and monotonicity between risk proxies and OOD accuracy, respectively. The values of both coefficients fall between -1 and 1. A coefficient being close to -1 or 1 indicates a robust negative or positive correlation. Conversely, a value of 0 denotes no correlation~\citep{cohen2009pearson}. Following \citep{deng2022strong},  we use top-1 classification accuracy as a metric to measure model generalization.

\subsection{Existing Risk Proxies as Baselines}
\label{Sec:baselines}
We evaluate the effectiveness of vicinal assessment in enhancing the following risk proxies in unsupervised generalization prediction. 
1) \emph{Average Confidence} (AC) \citep{tu2023assessing, hendrycks17baseline}. The mean of the softmax confidence for each samples on the test set.
2) \emph{Effective Insurance} (EI)~\citep{deng2022strong} is the multiplication between the confidence of the image and a different view of it (\eg, rotation) if the predicted class of them is the same. Otherwise, it is equal to zero for this sample.
3) \emph{Consistency Invariance} (CI)~\citep{aithal2021robustness}  is the predicted probability of the transformed view affecting the predicted class of the original image.
4) \emph{Difference of Confidence} (DoC)~\citep{guillory2021predicting} is obtained by using the accuracy on the held-out validation set to subtract the gap between the AC on the validation set and the AC on the test set.
5) \emph{Average Thresholded Confidence} (ATC)~\citep{Garg2022Leveraging} quantifies the proportion of samples that have a softmax confidence score exceeding the threshold learned from the validation set.



\subsection{Main Observations}

\begin{table*}[t]
\centering
\setlength{\tabcolsep}{4mm}{
    \caption{\textbf{Comparing vicinal risk proxies (VRP) and empirical risk proxies (ERP) on various test sets.} 
    For each test set, the first row shows ERP results, and the second row shows VRP results.
    $\gamma$ and $\rho$ represent the Pearson's coefficient and the Spearman's correlation coefficient, respectively. \textsuperscript{$\clubsuit$} means near IND test sets. 
    The figures in \textbf{\textcolor{blue}{blue bold}} (\textbf{\textcolor{red}{red bold}}) indicate that the correlation coefficient of VRP is \textbf{\textcolor{blue}{higher}} (\textbf{\textcolor{red}{lower}}) than that of ERP with statistical significance (p-value $< 0.05$) based on the two-sample t-test.
    Otherwise, their difference is not statistically significant.} 
    \vspace{-1mm}
    \label{tab:benchmark_testing}\
    \resizebox{1\textwidth}{!}{
    \begin{tabular}{c c cc cc cc cc cc}
            \toprule
            \multirow{2}{*}{\textbf{Train}} & \multirow{2}{*}{\textbf{Test}} & \multicolumn{2}{c}{EI} & \multicolumn{2}{c}{AC} & \multicolumn{2}{c}{CI} & \multicolumn{2}{c}{DoC} & \multicolumn{2}{c}{ATC}  \\
            \cmidrule(lr){3-4} \cmidrule(lr){5-6} \cmidrule(lr){7-8} \cmidrule(lr){9-10} \cmidrule(lr){11-12}
            & & $\gamma$ & $\rho$ & $\gamma$ & $\rho$ & $\gamma$ & $\rho$& $\gamma$ & $\rho$ & $\gamma$ & $\rho$ \\
            \midrule
            \multirow{12}{*}{ImageNet} & \multirow{2}{*}{ImageNet-A} & 0.882 & 0.645 & 0.581 & 0.464 & 0.856 & 0.617   & 0.877 & 0.761 & 0.851 & 0.436  \\
            & & \mygreen{\bf 0.900}  & \mygreen{\bf 0.692} & \mygreen{\bf 0.624} & \mygreen{\bf 0.503} & \mygreen{\bf 0.905} & \mygreen{\bf 0.722} & \mygreen{\bf 0.908} & \mygreen{\bf 0.797} & \mygreen{\bf 0.866} & \mygreen{\bf 0.481}  \\
	    \cmidrule{2-12}

            & \multirow{2}{*}{ImageNet-R} & 0.914 & 0.814 & 0.736 & 0.625 & 0.873 & 0.729  & 0.898 & 0.862 & 0.937 & 0.887  \\
            & & \mygreen{\bf 0.956}  & \mygreen{\bf 0.931} & \mygreen{\bf 0.818} & \mygreen{\bf 0.736} & \mygreen{\bf 0.931} & \mygreen{\bf0.854} & \mygreen{\bf 0.905} & \mygreen{\bf 0.894} & \mygreen{\bf 0.967} & \mygreen{\bf 0.946} \\
            \cmidrule{2-12}
            
            &\multirow{2}{*}{ImageNet-S} & 0.893 & 0.853 & 0.742 & 0.711 & 0.868 & 0.820  & 0.911 & 0.919 & 0.948 & 0.915 \\
            & & \mygreen{\bf 0.920}  & \mygreen{\bf 0.871} & \mygreen{\bf 0.763} & \mygreen{\bf 0.728} & \mygreen{\bf 0.878} & \mygreen{\bf 0.840} & \mygreen{\bf 0.926} & \mygreen{\bf 0.931} & \mygreen{\bf 0.954} & \mygreen{\bf 0.953} \\
            \cmidrule{2-12}
       
            & \multirow{2}{*}{ObjectNet} & 0.961 & 0.949 & 0.788 & 0.777 & 0.958 & 0.946 & 0.819 & 0.834 & 0.841 & 0.860 \\
            & & \mygreen{\bf 0.975}  & \mygreen{\bf 0.972} & \mygreen{\bf 0.838} & \mygreen{\bf 0.814} & \mygreen{\bf 0.969} & \mygreen{\bf 0.962} & \mygreen{\bf 0.849} & \mygreen{\bf 0.868} & \mygreen{\bf 0.857} & \mygreen{\bf 0.876} \\
            \cmidrule{2-12}
            
            & \multirow{2}{*}{ImageNet-Blur} & 0.870 & 0.831 & 0.711 & 0.730 & 0.824 & 0.793   & 0.781 & 0.776 & 0.882 & 0.867  \\
            & & \mygreen{\bf 0.907}  & \mygreen{\bf 0.857} & \mygreen{\bf 0.737} & \mygreen{\bf 0.741} & \mygreen{\bf 0.829} & \mygreen{\bf 0.802} & \mygreen{\bf 0.821} & \mygreen{\bf 0.821} & \mygreen{\bf 0.912} & \mygreen{\bf 0.890} \\
            \cmidrule{2-12}

            & \multirow{2}{*}{ImageNet-V2\textsuperscript{$\clubsuit$}} & 0.889 & 0.884 & 0.609 & 0.501 & 0.882 & 0.870 & 0.982 & 0.979 & 0.993 & 0.990 \\
            & & \mygreen{\bf 0.895} & 0.881 & 0.613 & 0.513 & \mygreen{\bf 0.886} & \mygreen{\bf 0.887} & \mygreen{\bf 0.990} & \mygreen{\bf 0.984} &  0.995 & 0.993 \\

            \midrule

            \multirow{4}{*}{CIFAR10} & \multirow{2}{*}{CINIC} & 0.913 & 0.936 & 0.978 & 0.887 & 0.834 & 0.876 & 0.985 & 0.953 & 0.983 & 0.937 \\
            & & \mygreen{\bf 0.954} & \mygreen{\bf 0.956} & 0.979 & \mygreen{\bf 0.889} & \mygreen{\bf 0.875} & \mygreen{\bf 0.929} & 0.985 & \mygreen{\bf 0.956} & 0.982 & \mygreen{\bf 0.942} \\
            \cmidrule{2-12}


            & \multirow{2}{*}{CIFAR10.1\textsuperscript{$\clubsuit$}} & 0.886 & 0.905 & 0.982 & 0.972 & 0.804 & 0.811 & 0.992 & 0.985 & 0.991 & 0.982 \\
            & & 0.883 & \red{\bf 0.886}  & 0.982 & 0.972 & \mygreen{\bf 0.813} & \mygreen{\bf 0.855} & 0.992 & 0.985 & 0.991 & 0.982 \\
            \midrule

            \multirow{2}{*}{iWildCam} & \multirow{2}{*}{iWildCam-OOD} & 0.337 & 0.362 & 0.635 & 0.445 & 0.268 & 0.258 & 0.547 & 0.532 & 0.509 & 0.526 \\
            & & \mygreen{\bf 0.402} & \mygreen{\bf 0.393} & \mygreen{\bf 0.655} & \mygreen{\bf 0.495} &  \red{\bf 0.208} & \red{\bf 0.180} & \mygreen{\bf 0.556} & \mygreen{\bf 0.592} &  \mygreen{\bf 0.518} &  \mygreen{\bf 0.595} \\

            \bottomrule
    \end{tabular}
    }
}
\end{table*}
\vspace{-1mm}

\begin{figure*}[t]%
\centering
\includegraphics[width=\textwidth]{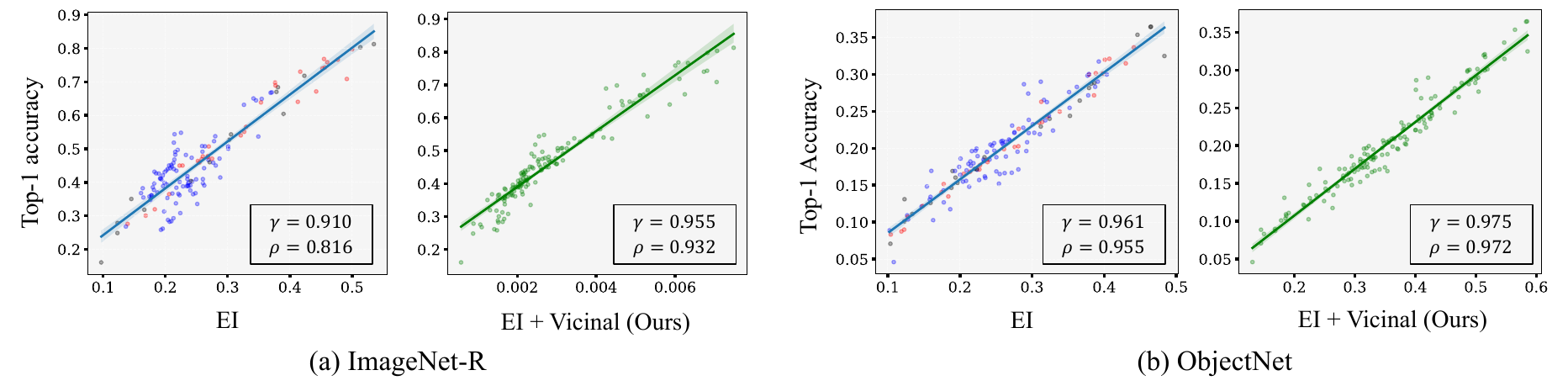}
\vspace{-2mm}
\caption{
\textbf{Correlation between Effective Invariance (EI) and Accuracy}: Each dot in the figures represents a model, and straight lines are fitted using robust linear regression \cite{Huber2011}. Blue dots represent the rectified score (VRP score) of these models, bringing their rank closer to the actual accuracy rank. Conversely, the rank of red models deviates further from the real accuracy when using the VRP paradigm. The rank of black models remains unchanged. The symbols $\rho$ and $\gamma$ have the same meaning as in Table \ref{tab:benchmark_testing}. The shaded region in each figure represents a 95\% confidence interval for the linear fit, calculated from 1,000 bootstrap samples. The VRP paradigm effectively rectifies the proxy score for the majority of models in both the ImageNet-R and ObjectNet datasets. Additional results for alternative risk proxies are presented in the supplementary material (Sec.~\ref{sec:append-data}).
}
\label{fig:scatter}
\vspace{-0.3cm}
\end{figure*}

\textbf{Statistically, vicinal risk scores better differentiate correct and incorrect model predictions.}
In addition to the example in Fig. \ref{fig:model_histogram}, we offer statistical evidence of our method's functionality. We use Gaussian kernel density estimation (KDE) to estimate the distributions of proxy scores for models making correct and incorrect predictions on each sample. Subsequently, we employ numerical integration to calculate their overlap, known as the overlap coefficient. Fig. \ref{fig:overlap_bar} presents the average coefficient across each test set. The results illustrate that vicinal assessment yields a lower overlap coefficient compared to baseline proxies, facilitating the distinction between correct and incorrect predictions at the individual sample level.

\textbf{Vicinal assessment consistently improves existing risk proxies on OOD test sets.} In Table \ref{tab:benchmark_testing}, we compare five existing risk proxies and their vicinal versions on OOD test sets. Each experiment is repeated three times to show statistical significance. We observe a consistent improvement in the strength of correlation between accuracy and risk proxy. For example, on ImageNet-A, vicinal assessment brings about 4.8\%, 3.9\%, 10.5\%, 3.6\%, and 4.5\% improvement in the Spearman's coefficient over EI, AC, CI, DoC, and ATC, respectively. These results indicate the effectiveness of the proposed method. The improvements can be illustrated by two examples in Fig, \ref{fig:scatter}, where the ranks of the majority models have been adjusted to be closer to the actual rank.

\textbf{Vicinal assessment is neither beneficial nor detrimental on near-OOD test sets.} When test data are near OOD or even IND, the use of vicinal assessment does not have noticeable performance improvement or compromise. For example, on ImageNet-V2, we observe slight improvement for EI, CI, DoC and ATC, and slight decrease for AC. On the CIFAR10.1 test set, similar observations are made. Further given its effectiveness in OOD scenarios, this allows for safe deployment of vicinal assessment in practice.

\begin{table*}[t]
\centering
\setlength{\tabcolsep}{4mm}{
\footnotesize
    \caption{\textbf{Comparison of variants of vicinal risk proxy.} 
    \textbf{(Top)}: various similarity metrics that can be used in Eq. \eqref{eq:vicinal_density_f}.  \textbf{(Bottom)}: various image transformations that can be used in Eq. \eqref{eq:vicinal_density_f}. Bold numbers denote the best one across compared settings.}
    \label{table:different_settings}
   
    \resizebox{1\textwidth}{!}{
    \begin{tabular}{l cc cc cc cc cc}
            \toprule
            \multirow{2}{*}{\textbf{Settings}} & \multicolumn{2}{c}{EI} & \multicolumn{2}{c}{AC} & \multicolumn{2}{c}{CI} & \multicolumn{2}{c}{DoC} & \multicolumn{2}{c}{ATC} \\
            \cmidrule(lr){2-3} \cmidrule(lr){4-5} \cmidrule(lr){6-7} \cmidrule(lr){8-9} \cmidrule(lr){10-11} 
            & $\gamma$ & $\rho$ & $\gamma$ & $\rho$ & $\gamma$ & $\rho$& $\gamma$ & $\rho$ & $\gamma$ & $\rho$ \\
            \midrule
            Random & 0.003 & 0.197 & 0.072 & 0.133 & 0.216 & 0.350 & 0.073 & 0.160 & 0.066 & 0.055  \\
            Equal & 0.883 & 0.659 & 0.554 & 0.446 & 0.857 & 0.624 & 0.864 & 0.757 & 0.837 & 0.424 \\
            Gaussian kernel & 0.876 & 0.675 & \textbf{0.611} & \textbf{0.498} & 0.887 & 0.706 & 0.887 & 0.797 & \textbf{0.881} & \textbf{0.532} \\
            Dot product & \textbf{0.903} & \textbf{0.713} & 0.605 & 0.489 & \textbf{0.903} & \textbf{0.731} & \textbf{0.901} & \textbf{0.801} & 0.867 & 0.490 \\
            \midrule
            \midrule
            None & 0.901& 0.705 & 0.564 & 0.480 & 0.881 & 0.669 & 0.907 & 0.806 &0.863 & 0.487 \\
            Grey-scale & 0.898 & 0.686 & \textbf{0.617} & \textbf{0.503} & 0.879 & 0.655 & 0.917 & \textbf{0.812} & \textbf{0.889} & \textbf{0.514} \\
            Color jitters & 0.897 & 0.674 & 0.632 & 0.512& 0.861 & 0.642 & \textbf{0.919} & 0.811 & 0.874 & 0.482 \\
            Rotation & 0.\textbf{903} & \textbf{0.713} & 0.605 & 0.489 & \textbf{0.903} & \textbf{0.731} & 0.901 & 0.801 & 0.867 & 0.490 \\

            \bottomrule

    \end{tabular}
    }
}
\end{table*}

\subsection{Further Analysis of Vicinal Risk Proxy}
\label{Sec:analysis_exp}

\textbf{Comparing different similarity measurements in Eq. \eqref{eq:vicinal_density_f}}. 
Dot product is used in Eq. \eqref{eq:vicinal_density_f} to compute similarity between a sample of interest and a sample in its vicinity. 
Here, we compare it with other options including random value (giving random similarity values), equal similarity (\ie, uniform vicinal distribution \citep{chapelle2000vicinal}), 
and the Gaussian kernel function (\ie, Gaussian vicinal distribution \citep{chapelle2000vicinal}). Results of ranking the 140 models on ImageNet-A are summarized in Table \ref{table:different_settings}. We find that dot product generally has similar performance with Gaussian similarity, and both are much better than random similarity and equal similarity. It illustrates the benefit of letting closer sample to contribute more to the score of the sample of interest. 

\begin{figure*}[t]
\centering
\includegraphics[width=\textwidth]{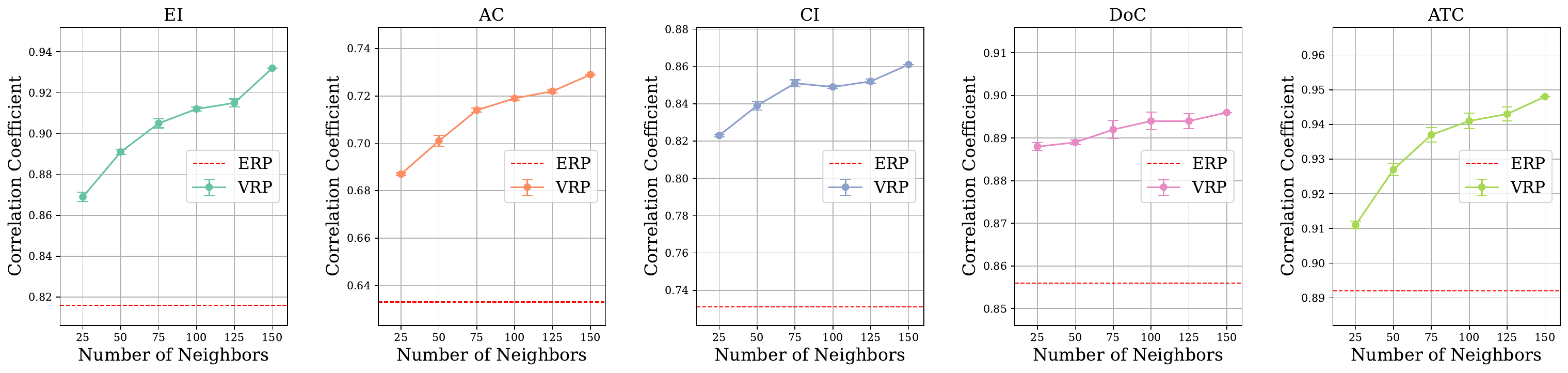}
\vspace{-5mm}
\caption{\textbf{Impact of the number of neighbors $m$ on the correlation between proxy scores and accuracy.} We use five existing proxies as baselines and report the mean and standard deviation for each data point. We observe that vicinal assessment is consistently beneficial under various $m$ values and yields stronger correlation with $m$ increases.}
\label{fig:neighbors_num}
\vspace{-0.3cm}
\end{figure*}

\textbf{Comparing different image transformations.} In Eq. \eqref{eq:vicinal_density_f}, we define the probability density function for vicinal distribution using transformed images. Here we compare the effectiveness of different transformations, and results on ImageNet-A are presented in Table \ref{table:different_settings}. We find that there is no significant performance difference between rotation, gray-scale transformation and color jitters.

\textbf{Impact of the number of neighbors.} {The number of neighbors $m$ is an important hyper-parameter used in Eq. \eqref{eq:vrp_approximation}. To evaluate system sensitivity to $m$, we experiment with ImageNet-R as OOD test set and five baseline risk proxies setting $m=25,50,75,100,125,150$\footnote{Because of the limited size of ImageNet-R, 150 is the maximum value we can set $m$ to.}.} 
We repeated each experiment three times and reported the mean and standard deviation.  From Fig. \ref{fig:neighbors_num} we have two observations. 
First, using more neighboring samples is generally beneficial, evidenced by the increasing correlation strength. It is probably because a larger $m$ allows for better approximation of the true vicinal distribution. We set $m=150$ by default. 
Second, when using fewer neighbors, \emph{e.g.,} $m=25$, vicinal assessment is still beneficial, yielding stronger correlation than existing risk proxies. 

We provide further analysis, including discussions on time complexity, evaluation of variants, and the effects of test set size in Appendix~\ref{sec:app-complex}, \ref{suppl:variants}, and~\ref{suppl:test-set-size}, respectively.

\section{Conclusion}
In this paper, we propose the vicinal assessment strategy to improve existing risk proxies computed based on a single test sample. We demonstrate that existing point-wise methods are prone to erroneous model responses, a problem that can be alleviated by considering the responses of adjacent test samples. Inspired by the philosophy of vicinal risk minimization, we design a vicinal risk proxy. We find that its computation on individual samples better differentiates models that make correct predictions from those that make incorrect ones. Therefore, when averaged across the test set, the vicinal risk proxy more accurately reflects the out-of-distribution (OOD) generalization ability of models. This main conclusion is verified through extensive experiments and further supported by analysis of its variants, sensitivity to key hyper-parameters, and application scope.

\newpage

{\small
\bibliographystyle{plainnat}
\bibliography{neurips_2024}
}

\clearpage

\newpage

\clearpage
\setcounter{page}{1}

\appendix
\renewcommand{\thefigure}{\Alph{section}\arabic{figure}}
\renewcommand{\thetable}{\Alph{section}\arabic{table}}
\setcounter{figure}{0}
\setcounter{table}{0}

\section{Experimental Setup}

In this section, we introduce how to access the models and datasets used in our paper.

\subsection{ImageNet Setup}
\label{sec:imagenet_setup}

\textbf{Models.}
Following the practice of \citet{deng2022strong}, we use the ImageNet models provided by PyTorch Image Models (timm) \citep{rw2019timm}. It provides models trained or fine-tuned on the ImageNet-1k training set \citep{deng2009imagenet}. We show the names of models used in our paper below:

\textit{`esmlp\_36\_224', 'cait\_s36\_384', 'cait\_s24\_224', 'convit\_base', 'convit\_tiny', 'twins\_pcpvt\_base', 'eca\_nfnet\_l1', 'xcit\_tiny\_24\_p8\_384\_dist',
'efficientnet\_b1', 'efficientnet\_b3' 'efficientnet\_b4', 'tf\_efficientnet\_b2',  'tf\_efficientnet\_lite1',  'convnext\_base',  'convnext\_small', 
'resnetrs350', 'pit\_xs\_distilled\_224', 'crossvit\_small\_240', 'botnet26t\_256', 'tinynet\_e', 'tinynet\_d', 'repvgg\_b2g4', 'mnasnet\_small', 'dla46x\_c', 'lcnet\_050', 'tv\_resnet34', 'tv\_resnet50', 'tv\_resnet101' 'tv\_resnet152' 'densenet121' 'inception\_v4' 'resnet26d' 'mobilenetv2\_140', 'hrnet\_w40',
'xception', 'xception41', 'resnet18', 'resnet34', 'seresnet50', 'mobilenetv2\_050', 'seresnet33ts', 'wide\_resnet50\_2', 'wide\_resnet101\_2', 'resnet18d', 
'hrnet\_w18\_small', 'gluon\_resnet152\_v1d', 'hrnet\_w48',  'hrnet\_w44',  'repvgg\_b2',  'densenet201',  'hrnet\_w18\_small',  'resnet101d', 'gluon\_resnet101\_v1d', 
'gluon\_resnet101\_v1s', 'gluon\_xception65', 'gluon\_seresnext50\_32x4d', 'gluon\_senet154', 'gluon\_inception\_v3', 'gluon\_resnet101\_v1c', 'tf\_inception\_v3', 
'tv\_densenet121', 'tv\_resnext50\_32x4d', 'repvgg\_b1g4', 'resnext26ts', 'ghostnet\_100', 'crossvit\_9\_240', 'deit\_base\_patch16\_384', 'rexnet\_150', 'rexnet\_130', 
'resnetrs50', 'resnet50d', 'resnet50', 'resnetv2\_50', 'resnetrs152', 'resnetrs101', 'dpn92', 'dpn98', 'dpn68', 'vgg19\_bn', 'vgg16\_bn', 'vgg13\_bn', 'vgg11\_bn', 
'vgg11', 'vgg11\_bn', 'vgg16', 'vgg19', 'swin\_small\_patch4\_window7\_224', 'swin\_base\_patch4\_window12\_384', 'deit\_base\_patch16\_224', 
'deit\_small\_distilled\_patch16\_224', 'densenet161', 'tf\_mobilenetv3\_large\_075', 'inception\_v3',
'ssl\_resnext101\_32x8d', 'ssl\_resnext101\_32x16d', 'swsl\_resnext101\_32x8d', 
'swsl\_resnext101\_32x16d', 'ssl\_resnext101\_32x4d', 'ssl\_resnext50\_32x4d', 'ssl\_resnet50', 
'swsl\_resnext101\_32x4d', 'swsl\_resnext50\_32x4d', 'swsl\_resnet50', 'tf\_efficientnet\_l2\_ns\_475', 
'tf\_efficientnet\_b7\_ns', 'tf\_efficientnet\_b6\_ns', 'tf\_efficientnet\_b4\_ns', 'tf\_efficientnet\_b5\_ns', 
'convnext\_xlarge\_384\_in22ft1k', 'convnext\_xlarge\_in22ft1k', 'convnext\_large\_384\_in22ft1k',
'convnext\_large\_in22ft1k', 'convnext\_base\_384\_in22ft1k', 'convnext\_base\_in22ft1k', 'resnetv2\_152x2\_bitm', 
'resnetv2\_152x4\_bitm', 'resnetv2\_50x1\_bitm', 'resmlp\_big\_24\_224\_in22ft1k', 'resmlp\_big\_24\_distilled\_224', 
'tf\_efficientnetv2\_s\_in21ft1k', 'tf\_efficientnetv2\_m\_in21ft1k', 'tf\_efficientnetv2\_l\_in21ft1k', 
'tf\_efficientnetv2\_xl\_in21ft1k', 'vit\_large\_patch16\_384', 'swin\_large\_patch4\_window12\_384', 
'beit\_large\_patch16\_512', 'beit\_large\_patch16\_384', 'beit\_large\_patch16\_224', 'beit\_base\_patch16\_384', 
'vit\_base\_patch16\_384', 'vit\_small\_r26\_s32\_384', 'vit\_tiny\_patch16\_384', 'vit\_large\_r50\_s32\_384', 
'mixer\_b16\_224\_miil' 'resmlp\_big\_24\_224', 'resnetv2\_50x1\_bit\_distilled', 'ig\_resnext101\_32x16d', 
'ig\_resnext101\_32x32d', 'ig\_resnext101\_32x8d', 'ig\_resnext101\_32x48d', 
'regnety\_016', 'regnety\_032'. 
}

\textbf{Datasets.}
We present the test sets employed in the main paper to evaluate the aforementioned ImageNet models. Datasets mentioned below can be accessed publicly via the provided links.

\noindent\emph{ImageNet-A(dversarial)} \citep{hendrycks2021nae}: \\\textcolor{blue}{https://github.com/hendrycks/natural-adv-examples}. \\
\emph{ImageNet-R(endition)} \citep{hendrycks2021many}: \\\textcolor{blue}{https://github.com/hendrycks/imagenet-r}. \\
\emph{ImageNet-Blur} \citep{hendrycks2019benchmarking}: \\\textcolor{blue}{https://github.com/hendrycks/robustness}.\\
\emph{ImageNet-S(ketch)} \citep{wang2019learning}: \\\textcolor{blue}{https://github.com/HaohanWang/ImageNet-Sketch}.\\
\emph{ImageNet-V2} \citep{recht2019imagenet}: \\\textcolor{blue}{https://github.com/modestyachts/ImageNetV2}.\\
\emph{ObjectNet} \citep{barbu2019objectnet}: \\\textcolor{blue}{https://objectnet.dev/download.html}.\\

\subsection{CIFAR10 Setup}

\textbf{Models.} We employ 101 models trained on the CIFAR10 training set.  Among them, 82 are trained based on the implementation from \textcolor{blue}{https://github.com/kuangliu/pytorch-cifar}, following the practice of \citet{deng2022strong}. These models vary in their architectures and number of training epochs. Specifically, the following architectures are used:

\textit{`DenseNet121', `DenseNet169', `DenseNet201', `DenseNet161', `densenet\_cifar', `DLA', `SimpleDLA', `DPN26', `DPN92', `EfficientNetB0', `GoogLeNet', `LeNet', `MobileNet', `MobileNetV2', `PNASNetA', `PNASNetB', `PreActResNet18', `PreActResNet34', `PreActResNet50', `PreActResNet101', `PreActResNet152', `RegNetX\_200MF', `RegNetX\_400MF', `RegNetY\_400MF', `ResNet18', `ResNet34', `ResNet50', `ResNet101', `ResNet152', 
`ResNeXt29\_2x64d', `ResNeXt29\_4x64d', `ResNeXt29\_8x64d', `ResNeXt29\_32x4d', `SENet18', `ShuffleNetG2', `ShuffleNetG3', `ShuffleNetV2', `VGG11', `VGG13', `VGG16', `VGG19'.
}

For each architecture, we train two variants with 30 and 50 training epochs, respectively.

The rest 19 models are download from \textcolor{blue}{https://github.com/chenyaofo/pytorch-cifar-models}. Names of these models are listed below:

\textit{`cifar10\_mobilenetv2\_x0\_5', `cifar10\_mobilenetv2\_x0\_75', `cifar10\_mobilenetv2\_x1\_0', `cifar10\_mobilenetv2\_x1\_4', `cifar10\_repvgg\_a0', `cifar10\_repvgg\_a1',  `cifar10\_repvgg\_a2', `cifar10\_resnet20', `cifar10\_resnet32', `cifar10\_resnet44', `cifar10\_resnet56', `cifar10\_shufflenetv2\_x0\_5', `cifar10\_shufflenetv2\_x1\_0', 
`cifar10\_shufflenetv2\_x1\_5', `cifar10\_shufflenetv2\_x2\_0', `cifar10\_vgg11\_bn', `cifar10\_vgg13\_bn', `cifar10\_vgg16\_bn', `cifar10\_vgg19\_bn'}.

\textbf{Datasets.} Datasets used in the CIFAR10 setup can be found through the following links.

\noindent\emph{CIFAR10} \citep{krizhevsky2009learning}: \\\textcolor{blue}{https://www.cs.toronto.edu/ kriz/cifar.html}. \\
\emph{CIFAR10.1} \citep{recht2018cifar10.1}: \\\textcolor{blue}{https://github.com/modestyachts/CIFAR-10.1}. \\
\emph{CINIC} \citep{darlow2018cinic}: \\\textcolor{blue}{https://github.com/BayesWatch/cinic-10}. \\

\subsection{iWildCam Setup}

\textbf{Models.} We access 34 models trained on the iWildCam training set from the official implementation of the iWildCam benchmark (\textcolor{blue}{https://worksheets.codalab.org/worksheets/0x52cea\\64d1d3f4fa89de326b4e31aa50a}). All models use Resnet50 \citep{he2016deep} as the backbone but differ in training algorithms, learning rate, weight decay, \textit{etc}. Their identification names are provided below. 

\textit{`iwildcam\_afn\_extraunlabeled\_tune0', 'iwildcam\_dann\_coarse\_extraunlabeled\_tune0', `iwildcam\_deepcoral\_coarse\_extraunlabeled\_tune0', `iwildcam\_deepcoral\_coarse\_singlepass\_extraun-labeled\_tune0', `iwildcam\_deepCORAL\_seed0', `iwildcam\_deepCORAL\_seed1', `iwildcam\_deepCORAL\_seed2'  `iwildcam\_deepCORAL\_tune', `iwildcam\_ermaugment\_tune0', `iwildcam\_ermoracle\_extraunlabeled\_tune0', `iwildcam\_erm\_seed0', `iwildcam\_erm\_seed1', 
`iwildcam\_erm\_seed2', `iwildcam\_erm\_tune0', `iwildcam\_erm\_tuneA\_seed0', `iwildcam\_erm\_tuneB\_seed0', `iwildcam\_erm\_tuneC\_seed0', `iwildcam\_erm\_tuneD\_seed0', 
`iwildcam\_erm\_tuneE\_seed0', `iwildcam\_erm\_tuneF\_seed0', `iwildcam\_erm\_tuneG\_seed0', `iwildcam\_erm\_tuneH\_seed0', `iwildcam\_fixmatch\_extraunlabeled\_tune0',  `iwildcam\_groupDRO\_seed0', `iwildcam\_groupDRO\_seed1', `iwildcam\_groupDRO\_seed2', `iwildcam\_irm\_seed0', `iwildcam\_irm\_seed1', `iwildcam\_irm\_seed2', `iwildcam\_irm\_tune',  `iwildcam\_noisystudent\_extraunlabeled\_seed0', `iwildcam\_pseudolabel\_extraunlabeled\_tune0', `iwildcam\_swav30\_ermaugment\_seed0'}.

\textbf{Dataset.}
\emph{iWildCam-OOD} \citep{beery2020iwildcam} can be download from the the official guidence: \textcolor{blue}{https://github.com/p-lambda/wilds/}.


\begin{figure*}[t]%
\centering
\includegraphics[width=\textwidth]{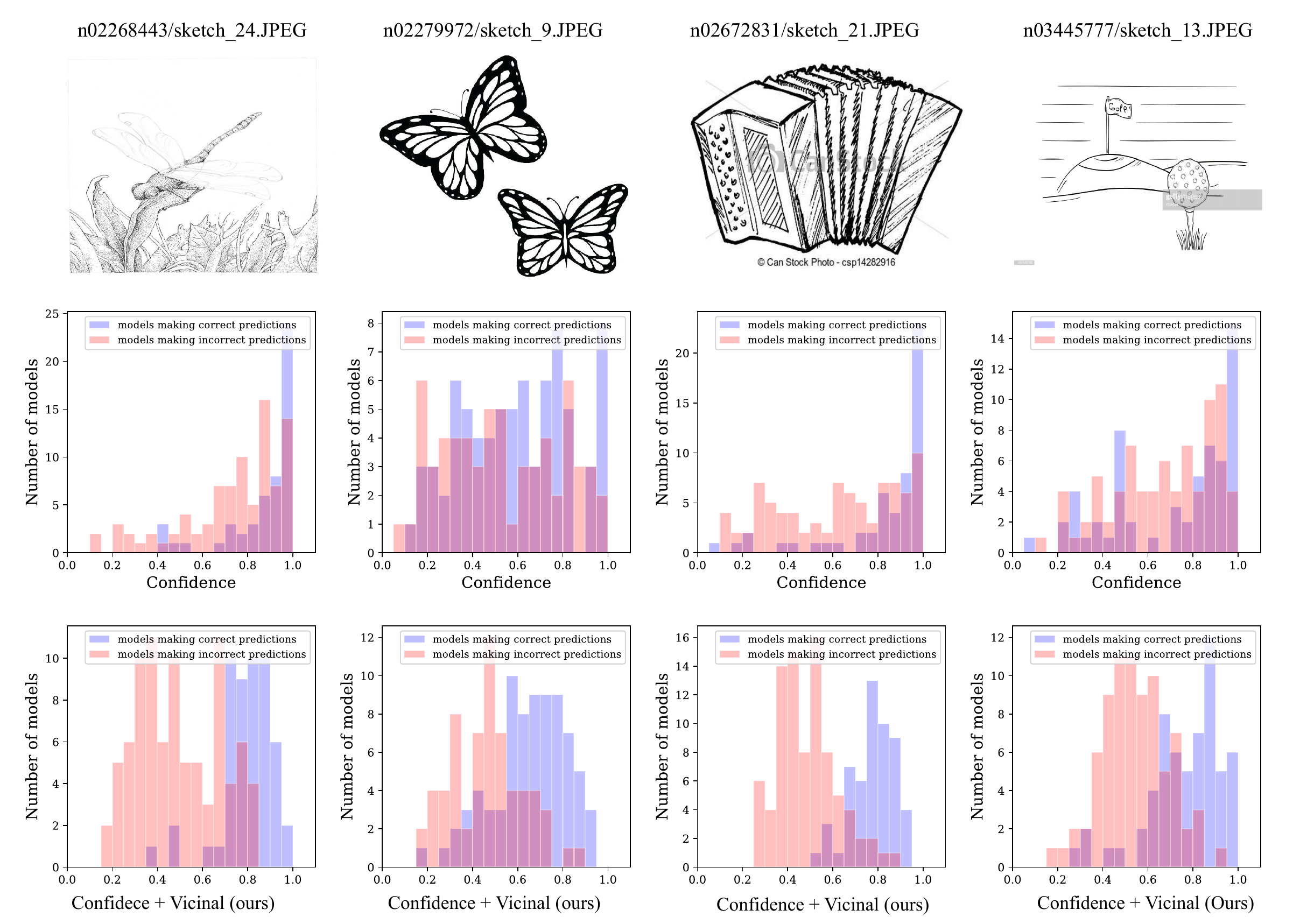}
\vspace{1mm}
\caption{\textbf{Illustration of working mechanism of vicinal assessment on individual samples 
from ImageNet-S.} Four test samples are shown in the top row. The AC method is used as baseline proxy. Score distribution of 140 models trained on the ImageNet training set are drawn below. Notations have the same meaning as Fig. 2 in the main paper.}
\vspace{1mm}
\label{fig:model_histogram_AC}
\end{figure*}

\begin{figure*}[t]%
\centering
\includegraphics[width=\textwidth]{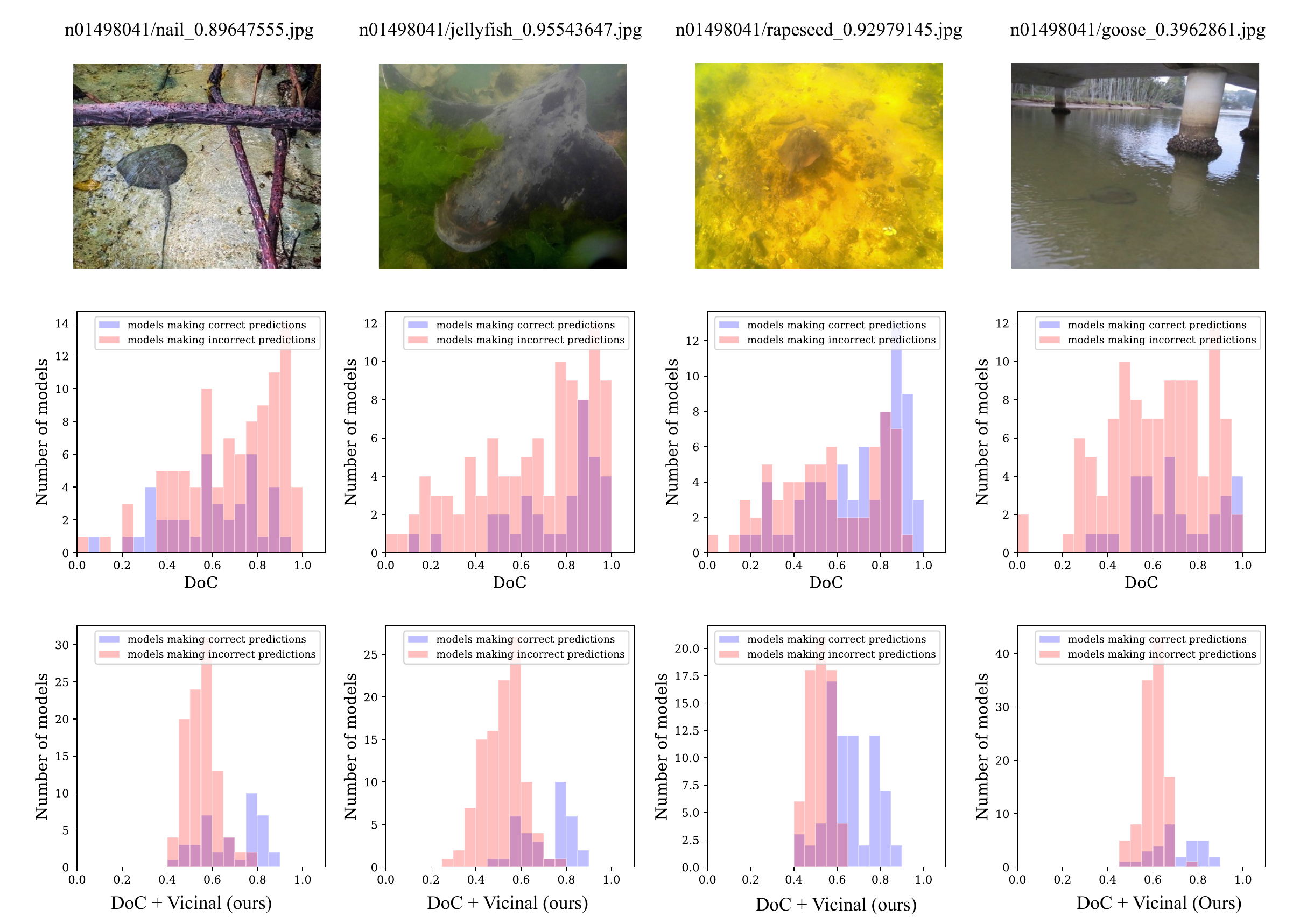}
\vspace{-5mm}
\caption{\textbf{Illustration of working mechanism of vicinal assessment on individual samples 
from ImageNet-A.}  DoC is used as baseline proxy. Other notations have the same meaning as Fig. \ref{fig:model_histogram_AC}.} 
\label{fig:model_histogram_DoC}
\end{figure*}

\begin{figure*}[t]%
\centering
\includegraphics[width=\textwidth]{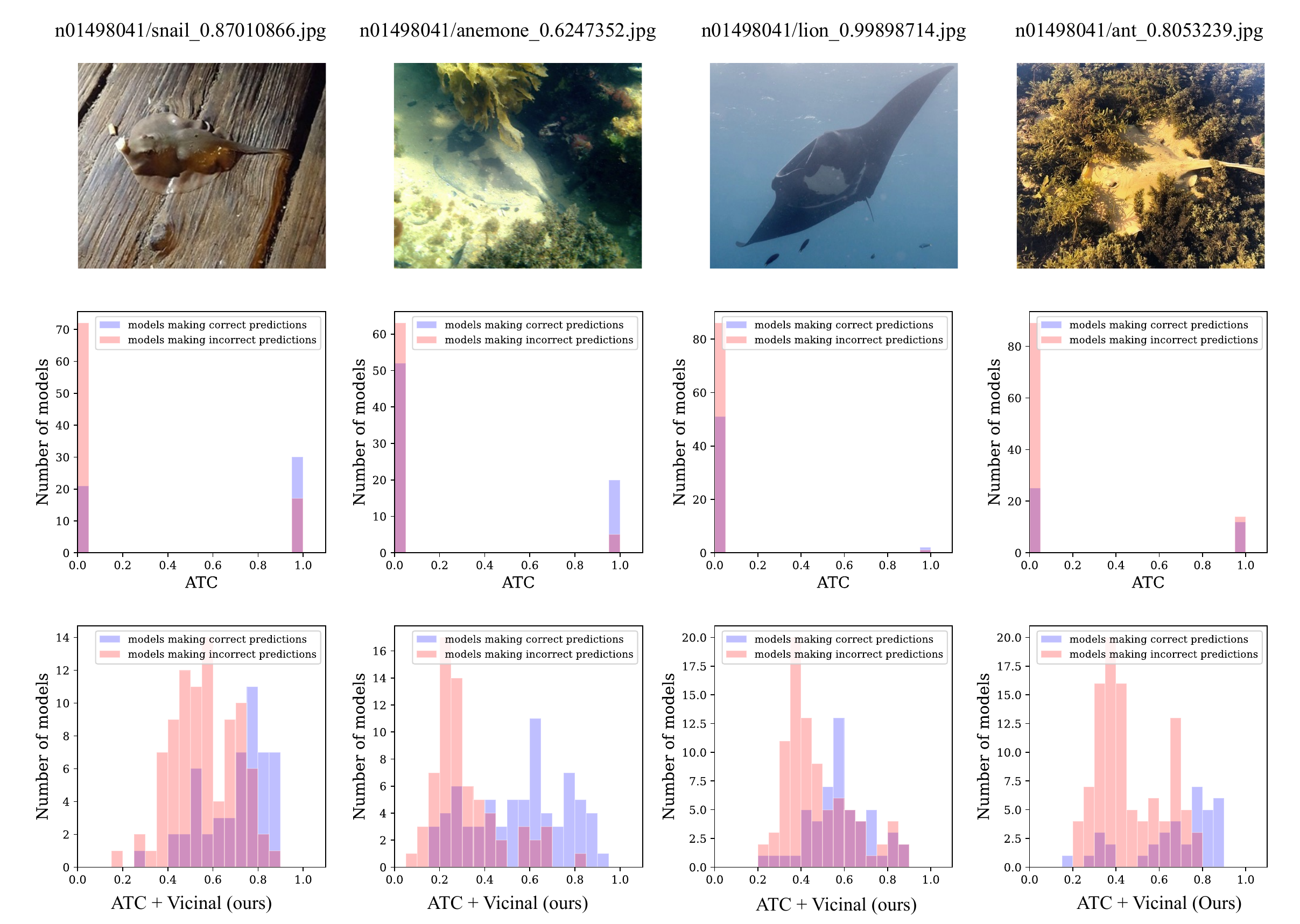}
\vspace{-5mm}
\caption{\textbf{Illustration of working mechanism of vicinal assessment on individual samples 
from ImageNet-R.} ATC is used as baseline proxy. Other notations have the same meaning as Fig. \ref{fig:model_histogram_AC}.}
\label{fig:model_histogram_ATC}
\end{figure*}

\section{Working Mechanism from Individual Test Samples to A Whole Test Set}

As discussed in the working mechanism part (see Section 4.2 of the main paper), we attribute the effectiveness of our method to its ability to distinguish correct and incorrect model predictions on an individual sample level, and thus better separability of models with different OOD performance.   In addition to the example shown in Fig. 2 of the main paper, we provide more examples to demonstrate the working mechanism on individual test samples in Fig. \ref{fig:model_histogram_AC}, Fig. \ref{fig:model_histogram_DoC}, and Fig \ref{fig:model_histogram_ATC}. We clearly observe that the distributions of model risk estimates of incorrect and correct predictions are generally more spreadable by using proposed vicinal assessment. These examples further showcase the working mechanism, where samples in the vicinity effectively rectify erroneous risk estimates, so that risk estimates of individual samples better differentiate models making corrent and incorrect predictions. The advantage of our method exhibited on individual test samples can guarantee a successful model ranking across the entire dataset. Here is the derivation.


\textbf{Definitions.} There are two models, $f_a,$ and $f_b$, to test. We denote each test sample in the test set as $\bm{x}_i, i \in \{1, 2, \ldots, n\}$. Given an baseline empirical risk proxy $\widehat{R_e}$ (\eg, AC, EI, and DoC), the empirical risk score of model $f_a$ on the test sample $\bm{x}_i$ is written as $\widehat{R_e}(f_a, \bm{x}_i)$. Similarly, the vicinal risk score of $f_a$ in the vicinity of $\bm{x}_i$ can be written as $\widehat{R_v}(f_a, \bm{x}_i)$. We assume $f_a$ has higher accuracy than $f_b$ and the risk proxy score is positively related to the model accuracy. 
Here, We define $p_e^i$ as the probability of the event that the empirical risk score for $f_a$ on the test sample $\bm{x}_i$ is higher than that for $f_b$, where
$$p_e^i = P((\widehat{R_e}(f_a, \bm{x}_i)-\widehat{R_e}(f_b, \bm{x}_i))>0).$$
Similarly, we have
$$p_v^i = P((\widehat{R_v}(f_a, \bm{x}_i)-\widehat{R_v}(f_b, \bm{x}_i))>0),$$
which means that the vicinal risk score for $f_a$ on the test sample $\bm{x}_i$ is higher than that for $f_b$.

\textbf{Working mechanism.} Given the preliminary assumption that the vicinal risk proxy $\widehat{R_v}$ enables the score distributions of correct and incorrect model predictions for each test sample to be more separable than the empirical risk proxy $\widehat{R_e}$(shown by our experiments in Fig. 3 of the main paper), there are some arbitrary samples $\{\bm{x}_i\}, i\in \{1, 2, \ldots, m\}, m\leq n$, where 
$$p_v^i > p_e^i, \forall i \in \{1, 2, \ldots, m\}.$$
In other words,
\begin{align*}
& P((\widehat{R_v}(f_a, \bm{x}_i)-\widehat{R_v}(f_b, \bm{x}_i))>0) \\
 &> P((\widehat{R_e}(f_a, \bm{x}_i)-\widehat{R_e}(f_b, \bm{x}_i))>0), \forall i \in \{1, 2, \ldots, m\}.
\end{align*}
Therefore, we have the following probability inequality:
\begin{align*}
& P\left(\left(\frac{1}{m}\sum_{i}^{m}\widehat{R_v}(f_a, \bm{x}_i)-\frac{1}{m}\sum_{i}^{m}\widehat{R_v}(f_b, \bm{x}_i)\right)>0\right) > \\
& P\left(\left(\frac{1}{m}\sum_{i}^{m}\widehat{R_e}(f_a, \bm{x}_i)-\frac{1}{m}\sum_{i}^{m}\widehat{R_e}(f_b, \bm{x}_i)\right)>0\right).
\end{align*}
According to Eq. 10 in the main paper,  the above inequality can be rewritten as
$$P((\widehat{R_v(f_a)}-\widehat{R_v(f_b)})>0) > P((\widehat{R_e(f_a)}-\widehat{R_e(f_b)})>0),$$
where $\widehat{R_v(f_a)}$ and $\widehat{R_v(f_b)}$ are the vicinal score and the empirical risk score for model $f_a$, respectively. It means that our method has the higher probability to successfully rank models $f_a$ and $f_b$.

\section{Time Complexity}
\label{sec:app-complex}

The computational complexity of our method is $O(nm)$, where $n$ is test set size, and $m$ is the number of neighboring samples used for VRP computation. In our algorithm, neighoring samples are those sharing the same predicted label as the sample of interest, so looking for neighboring samples does not require a search process. In our implementation, $m$ is a hyperparameter that can be as small as 25 or 50 (see Fig. 6) to yield improvement, while a test sample may have 150 neighbors.
So the computational complexity is much less than $O(n^2)$. 

We would like to present the experimental results of time consumption under different dataset sizes here. In Table \ref{tab:running time}, we provide the running time of evaluating the model `vit\_large\_r50\_s32\_384' on the ImageNet-R dataset. We used NVIDIA28
V100 with 4 × GPU.

\begin{table*}[t]
\centering
\caption{Running time (seconds) vs. test set size (\red{the whole test set}) for average confidence (AC) and our method (AC+VRP). The last row presents the runtime increase caused by applying VRP to AC. }
\setlength{\tabcolsep}{4.5mm}{
\begin{tabular}{lcccccc}
\toprule
Test set size & 5,000 & 10,000 & 15,000 & 20,000 & 25,000 & 30,000\\
\midrule
AC (sec.) & 1.75   & 1.88 \ &  1.95 \ & 1.99 \ & 2.14 \ & 2.22 \  \\
\midrule
AC + VRP (sec.) & 4.38   & 4.57 \ &  4.76 \ & 5.02 \ & 5.23 \ & 5.40 \  \\
\midrule
Added time (sec.) & 2.63  & 2.69 & 2.81  &  3.03 & 3.09 & 3.18 \\
\bottomrule
\end{tabular}
}
\label{tab:running time}
\end{table*}

Two main observations can be drawn from the above table. First, compared with the baseline average confidence (AC) method, applying VRP (AC+VRP) consumes an additional 2-4 seconds when the test set size is 30k.

Second, the runtime of VRP increases almost linearly with the test set size. For example, VRP runtime increases by 1.02 seconds and  when the test set size increases from 5k to 30k.

\section{More experiments on variants of vicinal risk proxy}
\label{suppl:variants}
We conduct experiments using different variants of the vicinal risk proxy on the ImageNet-R and ObjectNet datasets. The experimental results in Table ~\ref{table:different_settings_A} show similar observations to those in the main manuscript.

\begin{table*}[t]
\centering
\setlength{\tabcolsep}{2.5mm}{
\footnotesize
    \caption{\textbf{Comparison of variants of vicinal risk proxy on ImageNet-R (Top) and ObjectNet (Bottom).} 
    Notations follow Tab. 2 of the main submission.}
    \label{table:different_settings_A}
   
    \resizebox{0.98\textwidth}{!}{
    \begin{tabular}{l cc cc cc cc cc}
            \toprule
            \multirow{2}{*}{\textbf{Settings}} & \multicolumn{2}{c}{EI} & \multicolumn{2}{c}{AC} & \multicolumn{2}{c}{CI} & \multicolumn{2}{c}{DoC} & \multicolumn{2}{c}{ATC} \\
            \cmidrule(lr){2-3} \cmidrule(lr){4-5} \cmidrule(lr){6-7} \cmidrule(lr){8-9} \cmidrule(lr){10-11} 
            & $\gamma$ & $\rho$ & $\gamma$ & $\rho$ & $\gamma$ & $\rho$& $\gamma$ & $\rho$ & $\gamma$ & $\rho$ \\
            \midrule
            Random (to do) & 0.335 & 0.399 & 0.277 & 0.331 & 0.415 & 0.466 & 0.237 & 0.461 & 0.357 & 0.563  \\
            Equal & 0.910 & 0.816 & 0.738 & 0.633 & 0.868 & 0.731 & 0.897 & 0.856 & 0.935 & 0.892 \\
            Dot product & 0.955 & 0.932 & 0.817 & 0.729 & 0.925 & 0.861 & 0.909 & 0.896 & 0.966 & 0.948 \\
            \midrule
            None & 0.816 & 0.743 & 0.704 & 0.606 & 0.874 & 0.731 & 0.889 & 0.825 & 0.894 & 0.786 \\
            Grey-scale & 0.945 & 0.922 & 0.816 & \textbf{0.756} & \textbf{0.944} & \textbf{0.891} & 0.901 & \textbf{0.901} & \textbf{0.971} & \textbf{0.951}  \\
            Color jitters & 0.930 & 0.899 & 0.813 & 0.748 & 0.941 & 0.889 & 0.880 & 0.886 & 0.969 & 0.947 \\
            Rotation & \textbf{0.955} & \textbf{0.932} & \textbf{0.817} & 0.729 & 0.925 & 0.861 & \textbf{0.909} & 0.896 & 0.966 & 0.948 \\


            \midrule
            \midrule

            Random & 0.024 & 0.296 & 0.165 & 0.235 & 0.306 & 0.451 & 0.133 & 0.255 & 0.172 & 0.189  \\
            Equal & 0.894 & 0.851 & 0.748 & 0.714 & 0.865 & 0.819  & 0.913 & 0.921 & 0.950 & 0951 \\
            Dot product & \textbf{0.917}  & \textbf{0.871} & \textbf{0.763} & \textbf{0.729} & \textbf{0.875} & \textbf{0.837} & \textbf{0.927} & \textbf{0.932} &\textbf{0.955} & \textbf{0.955} \\
            \midrule
            None & 0.963 & 0.959 & 0.807 & 0.786 & 0.965 & 0.953 & \textbf{0.854} & \textbf{0.875} & 0.825 & 0.860 \\
            Grey-scale &  0.958 & 0.965 & 0.835 & \textbf{0.819} & 0.955 & 0.939 & 0.845 & 0.851 & 0.859 & 0.876 \\
            Color jitters & 0.955 & 0.937 & 0.833 & 0.804 & 0.948 & 0.924 & 0.844 & 0.862 & \textbf{0.864} & \textbf{0.880} \\
            Rotation & \textbf{0.975} & \textbf{0.972} & \textbf{0.838} & 0.814 & \textbf{0.969} & \textbf{0.963} & 0.849 & 0.868 & 0.857 & 0.876 \\

            \bottomrule

    \end{tabular}
    }
}
\end{table*}

\begin{table}[t]
    \centering
    \setlength{\tabcolsep}{1.5mm}{
    \caption{\textbf{Impact of the test set size.} We evaluate our method on the ImageNet-R set with different number of test samples.}
    \label{tab:dataset_size}
    \begin{tabular}{lcccccc}
					\hline	
				    \# samples &  3,000 & 6,000 & 12,000 & 18,000 & 24,000 \\
					\hline 
					DoC & 0.856 & 0.848 & 0.877 & 0.874 & 0.874  \\
				DoC + Ours & \textbf{0.873} & \textbf{0.874} & \textbf{0.892} & \textbf{0.895}  & \textbf{0.896}\\

					\hline 
				\end{tabular}
    }
    \vspace{-1mm}
\end{table}

\section{Impact of test set size} 
\label{suppl:test-set-size}
Table \ref{tab:dataset_size} presents our method under varying test set sizes using ImageNet-R as an OOD test set. We observe that the performance of all compared methods drops under smaller test sets. Nevertheless, the use of vicinal assessment consistently improves the correlation strength of the baselines under each test set size, demonstrating the effectiveness of our method. 

\section{Vicinal Assessment for Data-centric Unsupervised Evaluation} 
\label{sec:append-data}
As discussed in Section 4.4 of the main paper, vicinal assessment technically can be applied in data-centric unsupervised evaluation by measuring scores of a fixed model on various test sets. To evaluate this, we conduct experiments using two models, \emph{Efficientet-b2} and \emph{Inception-v4}, obtained from the model zoo in Section \ref{sec:imagenet_setup}, on 95 testing sets sourced from the test set pool, ImageNet-C \citep{hendrycks2019benchmarking}. The experimental results, showcasing the performance the vicinal method under different risk proxies, are presented in Table \ref{table:data-centric}.

\begin{table*}[t]
\centering
\setlength{\tabcolsep}{2.5mm}{
\footnotesize
    \caption{\textbf{Effectiveness of vicinal assessment in data-centric unsupervised evaluation.} 
    For each model, the \emph{first} row shows results of baseline risk proxies, while the \emph{second} row gives results of their vicinal assessments. $\gamma$ and $\rho$ have the same meaning as described in the main paper.}
    \label{table:data-centric}
   
    \resizebox{1\textwidth}{!}{
    \begin{tabular}{l cc cc cc cc cc}
            \toprule
            \multirow{2}{*}{\textbf{Model}} & \multicolumn{2}{c}{EI} & \multicolumn{2}{c}{AC} & \multicolumn{2}{c}{CI} & \multicolumn{2}{c}{DoC} & \multicolumn{2}{c}{ATC} \\
            \cmidrule(lr){2-3} \cmidrule(lr){4-5} \cmidrule(lr){6-7} \cmidrule(lr){8-9} \cmidrule(lr){10-11} 
            & $\gamma$ & $\rho$ & $\gamma$ & $\rho$ & $\gamma$ & $\rho$& $\gamma$ & $\rho$ & $\gamma$ & $\rho$ \\
            \midrule
            \multirow{2}{*}{Efficientnet-b2} & 0.932 & 0.976  & 0.987 & 0.993  & 0.919 & 0.965  & 0.987 & 0.992   & 0.933 & 0.939\\
             & 0.942 & 0.979  & 0.989 & 0.994  & 0.918 & 0.961  & 0.989 & 0.994 & 0.937 & 0.938  \\
             \midrule
              \multirow{2}{*}{Inception-V4} & 0.938 & 0.970  & 0.960 & 0.988  & 0.931 & 0.963  & 0.960 & 0.988   & 0.991 & 0.994\\
             & 0.942 & 0.973  & 0.964 & 0.990  & 0.928 & 0.960  & 0.964 & 0.990 & 0.993 & 0.995  \\

            \bottomrule

    \end{tabular}
    }
}
\end{table*}

Our main observation is that the correlation results obtained using vicinal samples are similar to those relying merely on individual samples. In other words, there is no noticeable improvements as opposed to those model-centric experiments in the main paper. In fact, as illustrated in Fig. 1-4, vicinal proxies are helpful for distinguishing models \textit{w.r.t} their OOD accuracy. Its limited capability in distinguishing hard datasets from easy ones  leads to the observations in Table \ref{table:data-centric}. 

\begin{figure*}[t]%
\centering
\includegraphics[width=\textwidth]{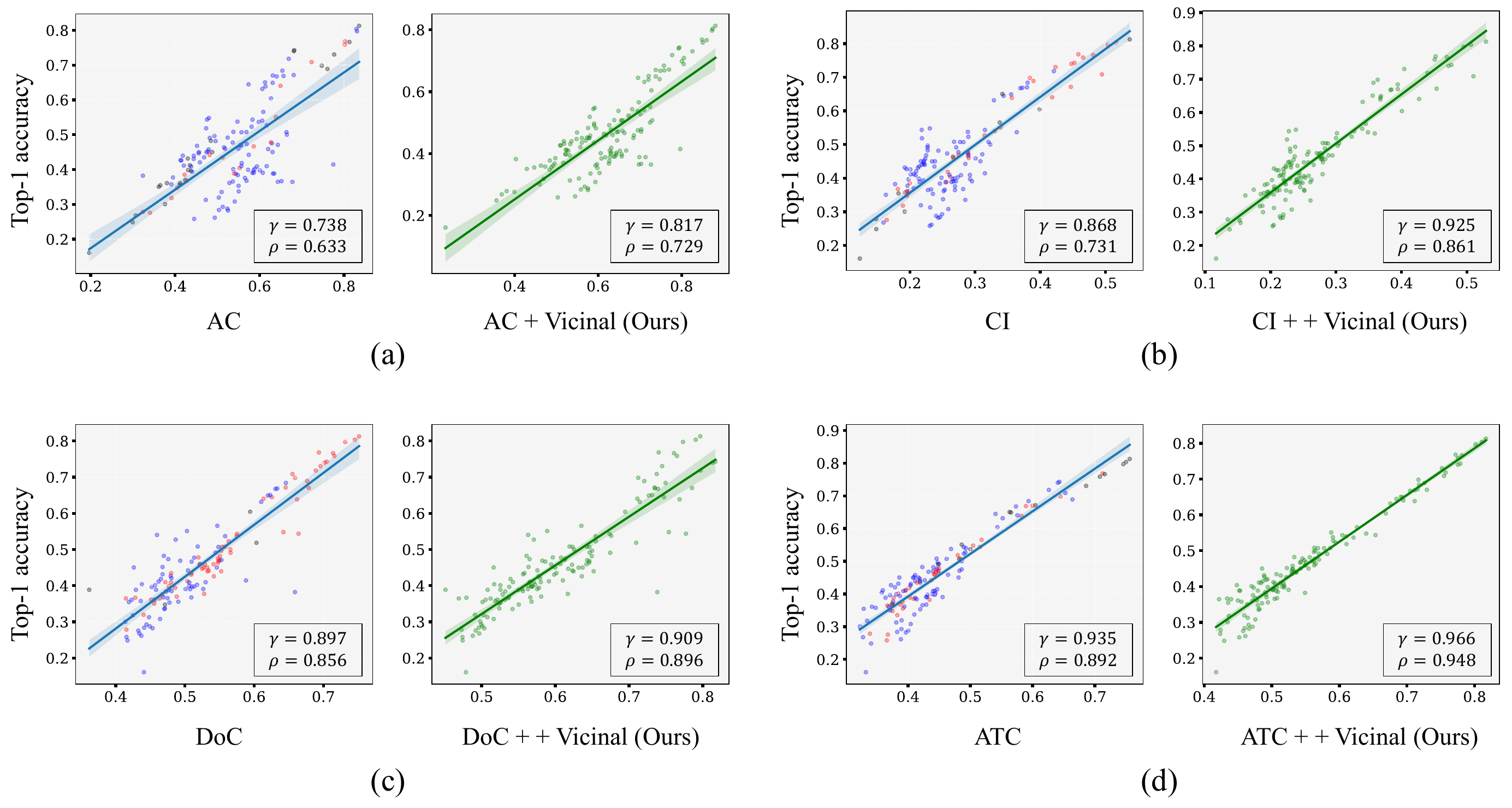}
\vspace{-5mm}
\caption{\textbf{Correlation between various risk proxies and accuracy on the ImgeNet-R dataset.} All notations in each figure have the same meanings as Fig. 5 of the main ppaer. We observe that proposed vicinal assessment effectively rectifies the risk estimates for the majority of models under various risk proxies.}
\label{fig:scatter_imagenet_r}
\end{figure*}

\begin{figure*}[t]%
\centering
\includegraphics[width=\textwidth]{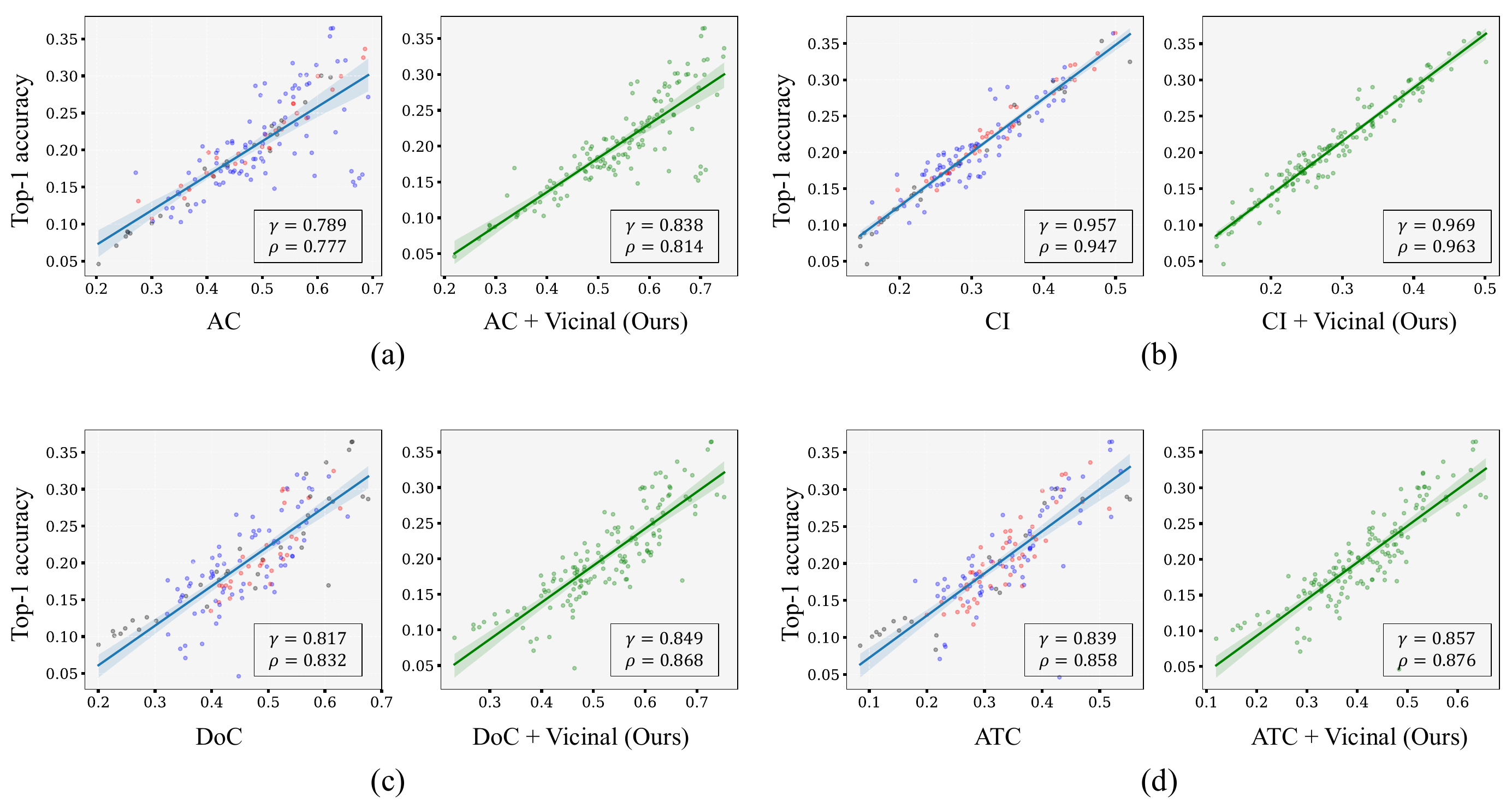}
\vspace{-5mm}
\caption{\textbf{Correlation between various risk proxies and accuracy on the ObjectNet dataset.} Notations in each figure have the same meanings as Fig. 5 of the main paper. Our observations are similar with those in Fig. \ref{fig:scatter_imagenet_r}.}
\label{fig:scatter_objectnet}
\end{figure*}

\begin{figure*}[t]%
\centering
\includegraphics[width=\textwidth]{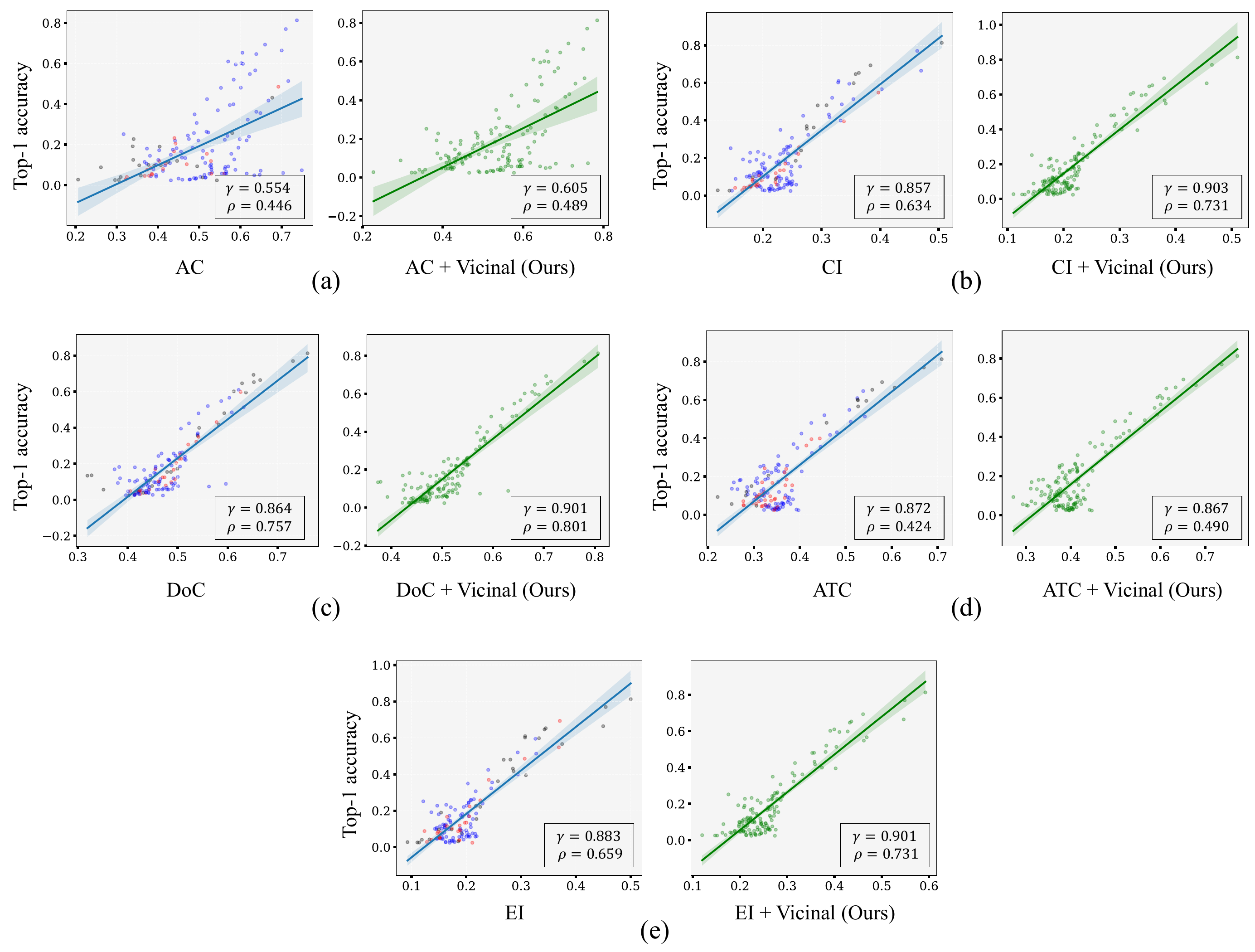}
\vspace{-5mm}
\caption{\textbf{Correlation between various risk proxies and accuracy on the ImageNet-A dataset.} Notations in each figure have the same meanings as Fig. 5 of the main paper. Our observations are similar with those in Fig. \ref{fig:scatter_imagenet_r} and Fig. \ref{fig:scatter_objectnet}.}
\label{fig:scatter_imagenet_a}
\end{figure*}

\section{More Visualizations of Improved Correlations}

In Fig. 5 of the main manuscript, we visualized correlations between effective invariance (EI) and accuracy and the improvement brought by vicinal proxies on the ImgeNet-R and ObjectNet datasets. Here, we present more visualizations, using AC, CI, DOC and ATC proxies. 
Results are shown in Fig. \ref{fig:scatter_imagenet_r}, Fig. \ref{fig:scatter_objectnet} and Fig. \ref{fig:scatter_imagenet_a}. We observe that generally the proposed vicinal assessment allows more models to get closer to the actual accuracy rank.

\section*{Limitation}

Our approach faces limitations in deployment environments with severely restricted computational resources, as it necessitates generating predictions from multiple images.

\section*{Impact statement}

This paper seeks to advance the field of machine learning. There are numerous potential societal benefits from our research, including the development of secure AI, explainable AI, and responsible AI.

\end{document}

%% file: math_commands.tex
\usepackage{amsmath,amsfonts,bm}

\def\eqref#1{equation~\ref{#1}}

\def\1{\bm{1}}

\DeclareMathAlphabet{\mathsfit}{\encodingdefault}{\sfdefault}{m}{sl}
\SetMathAlphabet{\mathsfit}{bold}{\encodingdefault}{\sfdefault}{bx}{n}

